\documentclass{article}

\usepackage[preprint]{neurips_2025}

\usepackage[utf8]{inputenc} 
\usepackage[T1]{fontenc}    
\usepackage{hyperref}       
\usepackage{url}            
\usepackage{booktabs}       
\usepackage{amsfonts}       
\usepackage{nicefrac}       
\usepackage{microtype}      
\usepackage{xcolor}         
\usepackage{graphicx}
\usepackage{booktabs}
\usepackage{bbm}
\usepackage{algorithm}
\usepackage{algorithmic}
\usepackage{multirow} 
\usepackage{soul}
\usepackage{url}
\usepackage{anyfontsize}
\usepackage{booktabs}
\usepackage{algorithm}
\usepackage{algorithmic}
\usepackage{amsfonts}
\usepackage{amssymb}
\usepackage{indentfirst}
\usepackage{bbding}
\usepackage{colortbl}
\usepackage{amsmath}
\usepackage{caption}
\usepackage{wrapfig} 
\usepackage{enumitem} 
\usepackage{changepage}  
\definecolor{customgreen}{rgb}{0.0, 0.7, 0.3}
\usepackage{amsmath}      
\usepackage{xcolor}       
\usepackage{tcolorbox}    
\usepackage{changepage}   
\definecolor{iccvblue}{rgb}{0.21,0.49,0.74}

\title{Zooming from Context to Cue: Hierarchical Preference Optimization for Multi-Image MLLMs}

\author{%
  Xudong Li$^{1}$ \quad Mengdan Zhang$^{2}$ \quad Peixian Chen$^{2}$ \quad Xiawu Zheng$^{1}$ \quad Yan Zhang$^{1}$  
  \\ \textbf{Jingyuan Zheng}$^{1}$ \quad \textbf{Yunhang Shen}$^{2}$ \quad \textbf{Ke Li}$^{2}$ \quad \textbf{Chaoyou Fu}$^{3}$ \quad \textbf{Xing Sun}$^{2}$ \quad \textbf{Rongrong Ji}$^{1}$
  \\
   $^{1}$ Key Laboratory of Multimedia Trusted Perception and Efficient Computing, \\
    Ministry of Education of China, Xiamen University, 361005, P.R. China\\
    $^{2}$ Tencent Youtu Lab \quad
    $^{3}$ Nanjing University\\
  \texttt \small{\{lxd761050753, zhangmengdanrz\}@gmail.com, \{zhengxiawu, rrji\}@xmu.edu.cn}\\
}

\begin{document}

\maketitle

\begin{abstract}

Multi-modal Large Language Models (MLLMs) excel at single-image tasks but struggle with multi-image understanding due to cross-modal misalignment, leading to hallucinations (context omission, conflation, and misinterpretation). Existing methods using Direct Preference Optimization (DPO) constrain optimization to a solitary image reference within the input sequence, neglecting holistic context modeling. We propose \textbf{C}ontext-to-\textbf{C}ue \textbf{D}irect \textbf{P}reference \textbf{O}ptimization~\textbf{(CcDPO)}, a multi-level preference optimization framework that enhances per-image perception in multi-image settings by zooming into visual clues---from sequential context to local details. It features: (i) \textit{Context-Level Optimization} : Re-evaluates cognitive biases underlying MLLMs' multi-image context comprehension and integrates a spectrum of low-cost global sequence preferences for bias mitigation. (ii) \textit{Needle-Level Optimization} : Directs attention to fine-grained visual details through region-targeted visual prompts and multimodal preference supervision. To support scalable optimization, we also construct \textbf{MultiScope-42k}, an automatically generated dataset with high-quality multi-level preference pairs. Experiments show that CcDPO significantly reduces hallucinations and yields consistent performance gains across general single- and multi-image tasks. Code will be available at \href{https://github.com/LXDxmu/CcDPO}{\textcolor{iccvblue}{link}}.

\end{abstract}    
\section{Introduction}
\vspace{-5pt}
Simultaneously understanding multiple images remains a fundamental yet underexplored challenge for Multi-modal Large Language Models (MLLMs)~\cite{ref1,ref2,ref3,ref4}. Despite MLLMs excelling in single-image tasks like visual question answering (VQA)~\cite{ref5,ref6,ref7}, code generation~\cite{ref8,ref9,ref10}, and storytelling~\cite{ref11, ref12}, and open-source models such as LLaVA~\cite{llava}, BLIP-2~\cite{ref13}, and InternVL~\cite{chen2024internvl} showing competitive results on benchmarks including VQAv2~\cite{ref14}, OKVQA~\cite{ref15}, and MMMU~\cite{ref16}, their capabilities in multi-image contexts are notably constrained. These models frequently struggle with tasks demanding cross-image comparison, spatial reasoning, or temporal alignment~\cite{ref17}, often resulting in hallucinations like context omission, conflation, and misinterpretation of local details. These deficiencies ultimately compromise model reliability. The root cause lies in the weak cross-modal alignment within MLLMs, which frequently fails to integrate visual and textual information coherently and comprehensively. This limitation becomes particularly pronounced in multi-image settings, where accurate reasoning demands: (i) Precise interpretation of intra-image regional details, and (ii) establishing meaningful inter-image connections through contextual integration.

To overcome these limitations, instruction tuning with multi-image supervision has been adopted in recent models such as Flamingo~\cite{ref2}, IDEFICS~\cite{laurenccon2024matters}, and Emu2~\cite{sun2024generative}. However, these approaches rely on large-scale annotated data, which is costly to construct due to the complexity of modeling inter-image relationships. As a lightweight alternative, \emph{Direct Preference Optimization (DPO)}~\cite{first-dpo} has emerged as a promising training paradigm, aligning model outputs with human preferences through pairwise supervision without requiring large-scale labeled data, significantly reducing reliance on costly annotations. Recent work has extended DPO to multimodal tasks~\cite{xie2024v,zhou2024aligning,huang2025vistadpo,zhang2024automated,ouali2024clip}, with MIA-DPO~\cite{mia-dpo} further pioneering its use in multi-image contexts by conditioning responses on a specific image through explicit query references (e.g., ``In Image 1, what is...?''), helping the model associate questions with the correct visual input.

\begin{figure*}[t]
  \centering
    \includegraphics[width=1\textwidth]{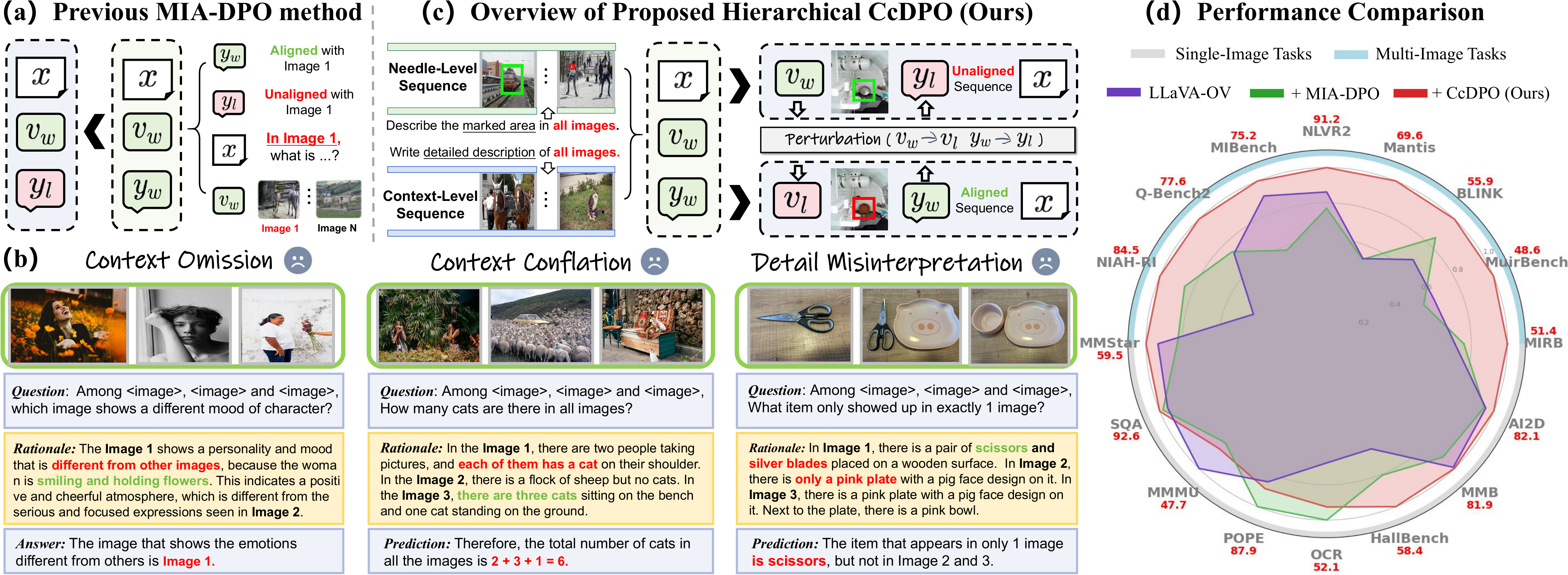}
  \caption{
(a) Prior multi-image DPO (e.g., MIA-DPO) is constrained by its reliance on predefined image references and text-only preferences, limiting holistic context modeling.
(b) These limitations commonly lead to failures such as Context Omission (ignoring relevant images), Context Conflation (misattributing content across images), and Detail Misinterpretation (misrepresenting fine-grained visual cues).
(c) CcDPO addresses these issues by hierarchically enhancing MLLMs' visual perception, from overall multi-image contexts to specific fine-grained details.
(d) Benchmark comparisons demonstrate CcDPO's improved reasoning capabilities on both multi-image and single-image tasks.}
  \label{fig1}
\vspace{-10pt}
\end{figure*}
While this explicit anchoring helps reduce ambiguity in how MLLMs interpret image references, its architecture remains fundamentally inadequate for addressing the intrinsic cross-modal misalignment within MLLMs. In particular, cross-image causal attention mechanisms exhibit heightened vulnerability to interference across multiple inputs~\cite{tian2025identifying,zhang2025cmmcot,wan2024look}. Consequently, MIA-DPO may struggle to autonomously perceive either the sequential visual context or local fine-grained details without explicit image references, resulting in a spectrum of multi-image hallucinations:
\textbf{Context Omission}: The model selectively ignores subsets of input images, generating responses based on incomplete sequences (Fig.~\ref{fig1}(b), left).
\textbf{Context Conflation}: The model erroneously attributes visual elements across images (e.g., describing a cat from Image 3 as appearing in Image 1; Fig.~\ref{fig1}(b), middle).
\textbf{Detail Misinterpretation}: Critical visual details in a certain image are either missed or misinterpreted. For example, in Fig.~\ref{fig1}(b) (right), without explicit image-specific instructions, the model fails to recognize the scissors in Image 2 and erroneously detects silver blades in Image 1.

To address these challenges, we propose \textbf{C}ontext-to-\textbf{C}ue \textbf{D}irect \textbf{P}reference \textbf{O}ptimization~\textbf{(CcDPO)}, a \textit{two-level preference optimization} framework that enhances MLLMs' capability to accurately perceive visual information across hierarchical levels—from sequential multi-image contexts to individual fine-grained details (as shown in Fig.~\ref{fig1}(c)). Specifically, it consist of two levels of alignment:

\noindent\textbf{(1) Context-Level Optimization:}
We leverage the low-cost preference construction inherent in multi-image captioning tasks by explicitly decomposing model responses into structured, per-image descriptions (e.g., ``For Image 1: <caption 1>'', ``For Image 2: <caption 2>''). This formulation enforces selective attention on individual images, ensuring contextual completeness while mitigating both inter-image interference and irrelevant visual content. To address \textbf{Context Omission} and \textbf{Context Conflation}, we introduce two perturbation techniques—\textit{sequence truncation} and \textit{content swapping}—into the captioning preference optimization process. By training the model to distinguish coherent contexts from disrupted ones, we promote holistic reasoning across the entire input sequence.

\noindent\textbf{(2) Needle-Level Optimization:}
To address \textbf{Detail Misinterpretation}, we propose a fine-grained preference learning strategy that sharpens the model's sensitivity to critical visual cues. Our approach integrates region-focused visual prompts into the preference data and employs DPO training to bias the model toward descriptions aligned with highlighted regions. This enhances the model's ability to detect, attend to, and describe salient visual elements across multiple images. Furthermore, inspired by~\cite{fu2025chip}, we incorporate vision contrastive preference supervision by constructing image pairs with varying alignment to reference descriptions. This encourages the model to refine its preference judgments on fine-grained visual cues within each image under contextual settings.

To support both optimization levels, we construct \textbf{MultiScope-42k}, a large-scale, automatically generated dataset. It comprises high-quality \textit{chosen} responses, synthesized from accurate image- and region-level descriptions, alongside \textit{rejected} responses generated through targeted perturbations at both contextual and local detail levels. This generation pipeline is fully automated, cost-effective, and scalable to diverse data sources.
Our main contributions are summarized as follows:
\vspace{-5pt}
\begin{itemize}
\item We pioneer the investigation of cognitive bias in multi-image comprehension for MLLMs, categorizing three prevalent hallucination types. To address these challenges, we propose \textbf{C}ontext-to-\textbf{C}ue \textbf{D}irect \textbf{P}reference \textbf{O}ptimization~\textbf{(CcDPO)}, an innovative two-level preference optimization framework that enhances per-image perception in multi-image settings by analyzing visual clues—from sequential context to local details.


\item We design a low-cost Context-Level Optimization mechanism, incorporating structured multi-image captioning preferences and targeted perturbation techniques to ensure MLLMs' comprehensive and consistent global context understanding. Complementarily, we develop a Needle-Level Optimization mechanism that enhances fine-grained visual acuity through the integration of region-focused visual prompts and vision contrastive preference signals.

\item We construct \textbf{MultiScope-42k}, a large-scale, high-quality dataset for two-level multi-image preference learning. The fully automatic generation pipeline is cost-effective and scalable across diverse data sources. After direct preference optimization on this dataset, our method significantly reduces hallucinations and achieves superior performance on multi-image tasks.
\end{itemize}

\vspace{-10pt}
\section{Related Work}
\vspace{-5pt}
\noindent \textbf{Multi-modal Large Language Models.} Recent advances in MLLMs~\cite{llava,llava-ov,qwen2vl,chen2024internvl} have combined powerful large language models (LLMs) with visual encoders via lightweight connectors, achieving impressive performance across dialogue~\cite{liu2024mmdu}, visual question answering~(VQA)~\cite{ref5}, and image captioning tasks~\cite{rohrbach2018object}. These models are typically trained on image-text pairs with instruction tuning, yielding strong single-image understanding. However, they remain prone to hallucinations~\cite{amber,HallBench,sun2023aligning}, especially in multi-image scenarios where accurate reasoning requires modeling not only individual images but also their cross-image relationships.
Recent studies aim to advance multi-image understanding by incorporating image-text interleaved data~\cite{laurenccon2023obelics,zhu2023multimodal} during model training. This approach helps develop capabilities such as image comparison~\cite{wang2024muirbench,kazemi2024remi}, cross-image association~\cite{jiang2024mantis,chen2023understanding}, and temporal reasoning~\cite{meng2024mmiu,li2024mvbench}. Nevertheless, instruction tuning with such data remains costly due to the need for complex, fine-grained annotations—an issue exacerbated in multi-image settings.

\noindent \textbf{Direct Preference Optimization.}
To align LLM outputs with human preferences, Reinforcement Learning from Human Feedback (RLHF)\cite{ouyang2022training,sun2023aligning,ref4} maximizes the preference gap between favored and disfavored responses using reward models. Direct Preference Optimization (DPO)\cite{first-dpo} offers a more efficient alternative by removing explicit reward modeling and reinforcement learning, streamlining preference alignment through supervised contrastive learning.
Recent work investigates DPO’s generalization and stability across tasks~\cite{zhang2024automated, wu2025symmetrical,wang2024enhancing,yuan2025tarsier2}, with multimodal extensions~\cite{fu2025chip, jiang2024modality,zhang2024direct,zhang2024automated,ouali2024clip} to reduce hallucinations and enhance vision-language grounding in single-image settings.
However, current language-based DPO methods often neglect visual details. To address this, vision contrastive DPO approaches either disrupt images~\cite{xie2024v,jiang2024modality} or highlight key visual tokens~\cite{cui2024fine,gu2024token}, enhancing preference learning but focusing mainly on single-image tasks. MIA-DPO~\cite{mia-dpo} pioneered DPO to multi-image settings by anchoring prompts to specific images, achieving promising results on relevant benchmarks. However, its reliance on predefined references limits holistic context modeling and autonomous cross-image reasoning.
In contrast, we propose CcDPO, which explicitly models global context and fine-grained visual cues to enhance multi-image reasoning.

\noindent \textbf{Visual Prompting for MLLMs.}
Visual prompting has been widely used in vision models~\cite{rao2022denseclip,kirillov2023segment,ravi2024sam}. Manually annotated points, boxes, or masks—often encoded by a separate prompt encoder—can guide the model to adjust segmentation granularity or select specific instances. More recently, MLLMs have shown the ability to interpret visual prompts directly embedded in the image without additional prompt encoders~\cite{shtedritski2023does,yao2024cpt,wu2024controlmllm}. Unlike prior work, our method actively integrates visual prompts into preference data and uses DPO training to encourage the model to prefer descriptions aligned with prompted regions, which enhances the model’s sensitivity to visually grounded information.
\begin{figure*}[t]
  \centering
    \includegraphics[width=1.0\textwidth]{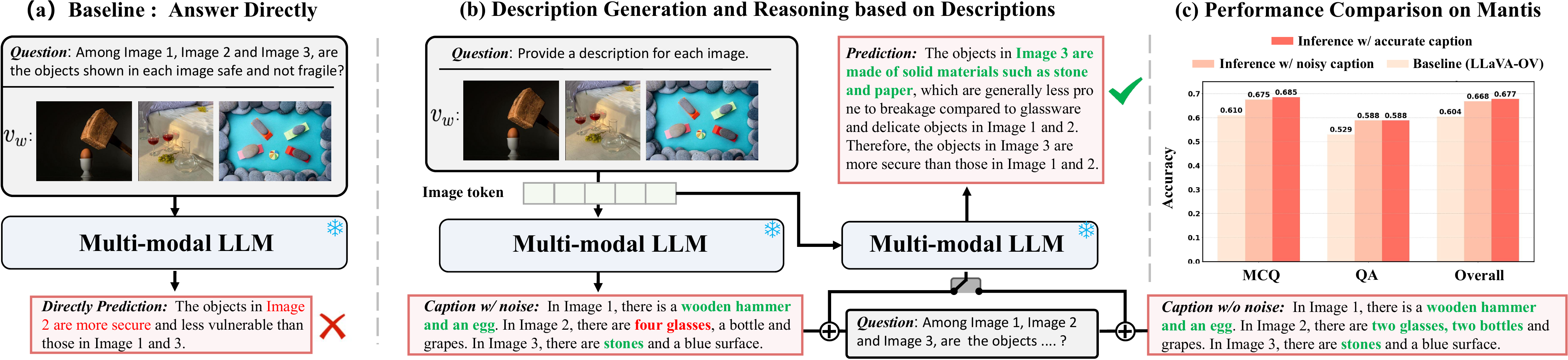}
  \caption{(a) Baseline: Direct inference without context as a condition. (b) Two-stage approach: Generating image captions, then reasoning over them. (c) Performance: Accurate caption understanding as context substantially improves VQA accuracy, with noisy captions also proving beneficial. This highlights deficient intrinsic captioning in MLLMs as a key bottleneck, motivating its enhancement.}
  \label{fig2}
\vspace{-15pt}
\end{figure*}
\vspace{-10pt}
\section{Exploring Cognitive Bias in Multi-Image Comprehension of MLLMs}
\label{sec3}
\vspace{-5pt}
\noindent \textbf{Limited Contextual Awareness.} 
We identify a fundamental limitation in current MLLMs: their impaired ability to perceive and integrate partial context information for coherent multi-image understanding severely degrades performance. To analyze this limitation, we examine the effect of context quality on multi-image reasoning, beginning by decoupling the multi-image VQA task into two stages (illustrated in Fig.~\ref{fig2}):
 (1) generating image-aligned descriptions (captions) of varying quality from sequential inputs, and (2) subsequently reasoning over these descriptions to derive an answer. We then conducted controlled experiments using variants of LLaVA-OV-7B~\cite{llava-ov}, evaluating its VQA performance when provided with no explicit context, noisy context, and accurate context augmentation.
As shown in Fig.~\ref{fig2} (c), our analysis indicates that providing accurate and explicit descriptions as context improves performance by about one point compared to noisy descriptions. More importantly, baseline models performing direct inference without such augmented context exhibit significantly degraded performance in multi-image settings ($\Delta$ Accuracy = 7.37). These findings confirm that deficient intrinsic multi-image captioning representation is a fundamental bottleneck for complex multi-image reasoning, motivating our direct efforts to enhance this skill.

\noindent \textbf{Multi-image Captioning Re-evaluation.}
We design a caption generation task as a proxy to systematically evaluate MLLMs' multi-image understanding, identifying three fundamental hallucination types—\textit{Context Omission}, \textit{Context Conflation}, and \textit{Detail Misinterpretation}—that critically degrade model performance.
To enable this evaluation, we construct \textbf{Context-AMBER-1K} by systematically concatenating images from the single-image AMBER dataset~\cite{amber} into sequences of two types: short-context sequences with 4 images and long-context sequences with 8 images. Each input is paired with the prompt: \textbf{\texttt{``Please sequentially describe each of the images shown above. Use the format: For Image *:<description>."}} The expected output format is: \textbf{\texttt{``For Image 1:<caption 1>,For Image 2:<caption 2>,..., For Image N:<caption N>."}}, ensuring comprehensive coverage of all input images. This controlled format allows precise identification of hallucination behaviors during multi-image comprehension.

We assess caption quality through four complementary metrics: (a) CHAIR~\cite{rohrbach2018object} measures object hallucination severity, (b) Response-level Hallucination Rate (Hal)~\cite{amber} quantifies incorrect descriptions, (c) Cognition-based Hallucination (Cog)~\cite{amber} detects reasoning errors, and (d) Sequence Coverage Rate (SCover) evaluates caption completeness across image sequences. Notably, for each input sequence, we compute hallucination scores (a), (b), and (c) for each image independently, then average these to get the sequence-level hallucination scores.

\noindent \textbf{Results.} As shown in Table~\ref{tab:hallucination}, even strong models like LLaVA-OV-7B exhibit significant hallucination rate increases in multi-image scenarios. For example, when input images grow from 4 to 8, \textit{Detail Misinterpretation hallucinations} become severe—the CHAIR score jumps from \textbf{10.2 to 50.6}, indicating a sharp decline in grounding accuracy. The sharp drop in the SCover score from \textbf{74.0\% to 10.3\%} also reveals \textit{context omission issues} within multi-image settings. Analysis of failure cases (Fig.~\ref{fig1}) reveals that \textit{Context Conflation} commonly occurs, critically degrading model performance.

\begin{figure*}[t]
  \centering
    \includegraphics[width=1\textwidth]{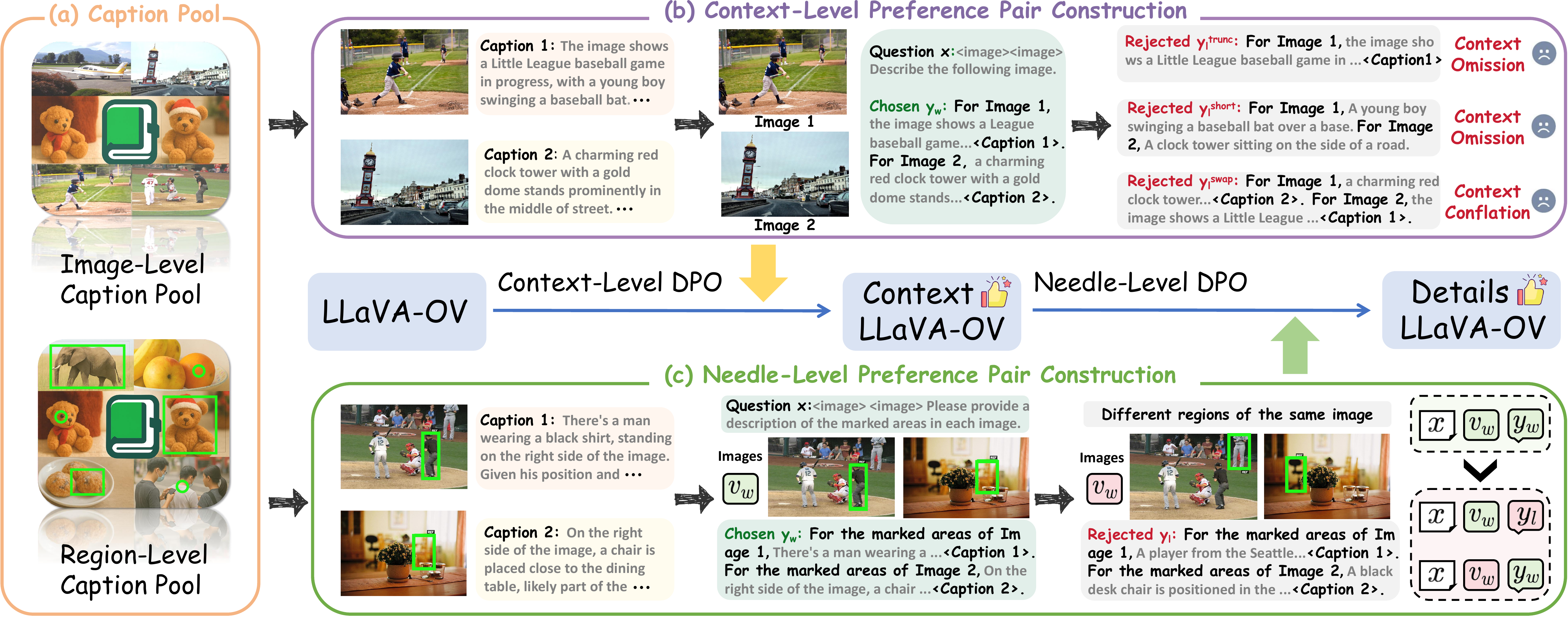}
  \caption{
  \textbf{Overview of CcDPO.}
  (a) Caption pools are built from LLaVA-23K~\cite{liu2023visual}, MDVP~\cite{lin2024draw}, and MVC~\cite{wu2025symmetrical} for image- and region-level supervision. 
  (b) Context-Level DPO aligns model outputs with complete, coherent image sequences and penalizes omissions, conflation, and misalignments. 
  (c) Needle-Level DPO incorporates visual prompts to enhance local detail understanding. chosen responses describe marked regions accurately, while rejected are drawn from mismatched regions. Both language-based and vision-contrastive preferences are used to sharpen fine-grained perception.
  }
  \label{fig3}
  \vspace{-10pt}
\end{figure*}
\vspace{-5pt}
\section{CcDPO: Context-to-Cue Direct Preference Optimization}
\vspace{-5pt}
As discussed in Sec.~\ref{sec3}, three fundamental hallucinations (context omission, context conflation, and detail misinterpretation) manifest as performance degradations in MLLMs' multi-image understanding. To address these challenges, we propose \textbf{C}ontext-to-\textbf{C}ue \textbf{D}irect \textbf{P}reference \textbf{O}ptimization~\textbf{(CcDPO)}, a hierarchical preference alignment framework that refines MLLMs at two levels (as shown in Fig.~\ref{fig3}): 

\noindent \textbullet\ \textbf{Context-Level Optimization:}  By contrasting complete and disrupted multi-image captions using language-based preference optimization, we enhance MLLMs' contextual understanding  by ensuring comprehensive integration of all relevant visual information across image sequences.


\noindent \textbullet\ \textbf{Needle-Level Optimization:} A hybrid preference optimization framework integrates two complementary objectives: (1) Contrasts captions that either align with or mismatch visually prompted regions using language-based preference optimization, and (2) Discriminates between images semantically matching or contradicting given captions using vision contrastive preference optimzation.
This dual approach trains the model to make preference judgments grounded in fine-grained visual details.


To support this, we construct \textbf{MultiScope-42k}, a large-scale preference dataset with automatically generated positive and perturbed response pairs at both levels.

\vspace{-5pt}
\subsection{Context-Level DPO with Language-Based Preference Optimization}
\label{sec:context}
\vspace{-5pt}
We propose a low-cost DPO mechanism for MLLMs that enforces coherent multi-image understanding to address context hallucinations (omission, conflation). Specifically, we reformulate the response generation task as a structured, per-image captioning problem. Each image in a sequence is described in an explicit format, generating the preferred response $\boldsymbol y_w$ as a coherent sequence of captions that reflect the content of each image.
 
\begin{adjustwidth}{0.0em}{0pt}  
$\boldsymbol y_w$ = [\texttt{For Image 1:} <caption 1>, \texttt{For Image 2:} <caption 2>,..., \texttt{For Image N:} <caption N>.]
\end{adjustwidth}
This encourages selective attention and attribution-aware generation. During training, we construct preference pairs where the \textbf{positive sample} is a coherent, full-sequence description $\boldsymbol y_w$, and the \textbf{negative sample} $\boldsymbol y_l$ is obtained from following two perturbation strategies:

\noindent \textbullet\ \textbf{Sequence Truncation}: simulates context omission by either removing captions entirely from one or more images (complete content omission) or replacing detailed captions with shorter versions (partial content omission), generating rejected responses \( y_l^{\text{trunc}} \) and \( y_l^{\text{short}} \). Complete omission disrupts sequence continuity, while partial omission results in sparse or incomplete sequence information:
\begin{adjustwidth}{0.5em}{0pt}
\centering
\( y_l^{\text{trunc}} \) = [\texttt{For Image 1:} <caption 1>, \texttt{For Image 3:} <caption 3>, \dots] \\
\( y_l^{\text{short}} \) = [\texttt{For Image 1:} <short caption 1>, \texttt{For Image 2:} <short caption 2>, \dots]
\end{adjustwidth}

\noindent \textbullet\ \textbf{Content Swapping}: simulates conflation by mismatching image indices and their descriptions, generating the rejected response $y_l$ by swapping the captions between different images. e.g.,
\begin{adjustwidth}{0.5em}{0pt}
\centering
$y_l^{\text{swap}}$ = [\texttt{For Image 1:} <caption 2>, \texttt{For Image 2:} <caption 1>, \dots]
\end{adjustwidth}

Given an instruction $\boldsymbol x$ and image sequence $\boldsymbol v_w$, we define the context-level DPO objective as:
\begin{equation}
\mathcal{L}_{\text{DPO}_t} = - \log \sigma \left( \beta \log \frac{\pi\theta(y_w \mid v_w, x)}{\pi{\text{ref}}(y_w \mid v_w, x)} - \beta \log \frac{\pi_\theta(y_l \mid v_w, x)}{\pi_{\text{ref}}(y_l \mid v_w, x)} \right), \quad y_l \in \{ y_l^{\text{trunc}}, y_l^{\text{short}}, y_l^{\text{swap}} \}
\label{tdpo}
\end{equation}
where $\pi_\theta$ is the target model and $\pi_{\text{ref}}$ is a frozen reference model. This objective teaches the model to prefer globally coherent, attribution-consistent responses over disrupted ones. In this stage, we use LLaVA-23K~\cite{liu2023visual} and coco~\cite{coco} as our detailed and brief context caption pool, respectively. 
\vspace{-5pt}
\subsection{Needle-Level DPO with Hybrid Visual-Language Optimization}
\label{sec:needle}
\vspace{-5pt}
Even when global context is preserved, MLLMs often fail to identify or attend to salient visual elements---e.g., missing objects, actions, or attributes. This leads to \textbf{detail misinterpretation}, which global response-based supervision alone cannot resolve. To address this, we introduce \textbf{needle-level optimization}, a fine-grained hybrid preference mechanism employing visual prompts and image-level perturbations to sharpen the model's focus on often-overlooked local visual cues.

\noindent\textbf{Language-based Preference Optimization.}
This stage leverages region-specific visual cues to guide the model’s preference judgments. We create preference pairs as follows:

\noindent\textbf{1) Chosen Responses ($\boldsymbol{y_r}$):} We integrate region-level visual prompts (e.g., bounding boxes, keypoints) into images $\boldsymbol{v}$ to highlight a target region $\boldsymbol{r}$, yielding $\boldsymbol{v_r}$. The model is trained to prefer the accurate description $\boldsymbol{y_r}$ of this specific region, directing its attention to critical visual elements.
\begin{adjustwidth}{0.0em}{0pt} 
\centering
$\boldsymbol y_r$ = [\texttt{For the marked area of Image 1:} <caption $r_1$>, \texttt{For the marked area of Image 2:} <caption $r_2$>..., \texttt{For the marked area of Image N:} <caption $r_N$>.]
\end{adjustwidth}

\noindent\textbf{2) Rejected Responses ($\boldsymbol{y_r'}$):} Descriptions \(\boldsymbol{y_{r'}}\) of regions \(\boldsymbol{r'}\) that are non-overlapping with \(\boldsymbol{r}\) within the same image serve as rejections. By learning to identify and reject such inaccuracies, the model can produce more precise and reliable captions.
\begin{adjustwidth}{0.0em}{0pt} 
\centering
$\boldsymbol y_r'$ = [\texttt{For the marked area of Image 1:} <caption $r'_1$>, \texttt{For the marked area of Image 2:} <caption $r'_2$>..., \texttt{For the marked area of Image N:} <caption $r'_N$>.]
\end{adjustwidth}
\noindent Given an instruction $\boldsymbol x$ and the image sequence $\boldsymbol{v_r}$, these pairs $(\boldsymbol x, \boldsymbol{v_r}, \boldsymbol{y_r}) \succ (\boldsymbol x, \boldsymbol{v_r}, \boldsymbol{y_r'})$ inform the language-based DPO objective (Eq.~\ref{tdpo}). We utilize MDVP~\cite{lin2024draw} for region-level caption pool.

\noindent\textbf{Vision Contrastive Preference Optimization.}
Inspired by~\cite{wu2025symmetrical,fu2025chip}, this stage further hones the model's visual discrimination. It trains the model by contrasting a single description $\boldsymbol{y_w}$ against two image inputs: $\boldsymbol{v_w}$, which correctly aligns with $\boldsymbol{y_w}$ (often focusing on a specific visual region), and $\boldsymbol{v_l}$, which is misaligned or visually contradicts $\boldsymbol{y_w}$. The objective combines two components:

\noindent\textbf{1) Focusing on Relevant Visuals ($\mathcal{L}_{\text{Focus}}$):} This rewards prioritizing details in the correctly aligned image $\boldsymbol{v_w}$ when generating $\boldsymbol{y_w}$, countering MLLMs' tendency to neglect visual content.
\begin{equation}
\mathcal{L}_{\text{Focus}}(v_w, y_w) = - \log \sigma \left( \beta_1 \log \frac{\pi_\theta(y_w \mid v_w, x)}{\pi_{\text{ref}}(y_w \mid v_w, x)} - \beta_1 \log \frac{\pi_\theta(y_w \mid x)}{\pi_{\text{ref}}(y_w \mid x)} \right),
\label{vdpo}
\end{equation}
\noindent\textbf{2) Rejecting Contradictory Visuals ($\mathcal{L}_{\text{Reject}}$):} This penalizes assigning high probability to $\boldsymbol{y_w}$ when conditioned on a contradictory image $\boldsymbol{x_l}$.
\begin{equation}
\mathcal{L}_{\text{Reject}}(v_l, y_w) = - \log \sigma \left( \beta_2 \log \frac{\pi_\theta(y_w \mid x)}{\pi_{\text{ref}}(y_w \mid x)} - \beta_2 \log \frac{\pi_\theta(y_w \mid v_l, x)}{\pi_{\text{ref}}(y_w \mid v_l, x)} \right),
\label{eq:reject_loss} 
\end{equation}
The combined vision contrastive DPO loss is $\mathcal{L}_{{\text{DPO}}_v}(v_w, y_w, v_l) = \mathcal{L}_{\text{Focus}}(v_w, y_w) + \mathcal{L}_{\text{Reject}}(v_l, y_w)$. This objective sharpens the model’s ability to distinguish fine-grained visual cues by rewarding focus on relevant details and penalizing attention to misleading content. We use the MVC~\cite{wu2025symmetrical} dataset as our region-level visual counterfactual caption pool.

\section{Experiments}
\subsection{Experimental Settings and Evaluation Benchmarks}
\noindent\textbf{Baselines.} We apply CcDPO to two different 7B-size MLLMs: Qwen2-VL~\cite{qwen2vl} and LLaVA-OV~\cite{llava-ov}. Due to differences in base models, preference data, and alignment strategies, direct comparisons with other LLMs are not possible. However, we provide the results for reference: LLaVA-1.5~\cite{li2024llava}, InternVL2-8B~\cite{chen2024expanding}, Mantis-Idefics~\cite{jiang2024mantis}, mPLUG-Owl3~\cite{ye2024mplug}, Idefics2-8B~\cite{laurenccon2024matters}, and Emu2-Chat~\cite{sun2024generative}.

\noindent\textbf{Implementation Details.}  
Our model undergoes a three-stage training process to better understand multi-image preferences at both broad (context) and detailed (needle) levels. \textbf{Stage 1} focuses on context-level alignment, where we fine-tune Qwen2-VL-7B and LLaVA-OV-7B for one epoch with learning rates of $5 \times 10^{-6}$ and $5 \times 10^{-5}$, respectively, using Eq.~\ref{tdpo}. \textbf{Stage 2} applies needle-level language-based DPO using Eq.~\ref{tdpo} to improve sensitivity to fine-grained visual cues with the same learning rate of $5 \times 10^{-5}$. We conduct Stage 1 and Stage 2 by using LoRA adaptation~\cite{hu2022lora} with rank $r=128$ for efficiency.
\textbf{Stage 3} performs vision contrastive DPO via full-parameter tuning for one epoch with a learning rate of $1 \times 10^{-6}$ using Eq.~\ref{vdpo}, strengthening the model's ability to distinguish preferred visual content. Following the setup in~\cite{mia-dpo}, we set the temperature parameter $\beta=\beta_1=\beta_2=0.1$ and the negative log-likelihood (NLL) loss coefficient $\gamma=0.1$. All training is conducted on 8 GPUs, each equipped with 90GB of memory.

\noindent\textbf{Evaluation Benchmarks.} We employ seven multi-image benchmarks—{MUIRBench}~\cite{wang2024muirbench}, {MIRB}~\cite{mirb}, BLINK~\cite{blink}, Mantis-Eval~\cite{jiang2024mantis}, NLVR2~\cite{nlvr2}, {Q-Bench2}~\cite{q-bench2}, and MIBench~\cite{liu2024mibench}—to holistically evaluate multi-image reasoning across four key dimensions: co-reference alignment, fine-grained comparison, contextual reasoning, and temporal understanding. Complementing these, eight representative single-image benchmarks assess specific multimodal capabilities: (1) Academic/Scientific Reasoning: {MMMU}~\cite{yue2024mmmu}, {MMStar}~\cite{mmstar}, {ScienceQA}~\cite{sqa}, (2) Diagram Understanding: {AI2D}~\cite{ai2d}, (3) Robustness against hallucinations: {POPE}~\cite{pope}, {HallBench}~\cite{HallBench}, (4) General Multimodal Abilities: {MMBench}~\cite{liu2024mmbench}, (5) Text Recognition: {OCRBench}~\cite{liu2024ocrbench}. This comprehensive evaluation suite demonstrates our method's strengths in both holistic understanding and fine-grained visual grounding across single-image and multi-image general tasks.

\begin{table*}[t]
\centering
\renewcommand{\arraystretch}{1.0}
\setlength{\tabcolsep}{3.5pt}
\caption{\textbf{Main results on general multi-image benchmarks.} We compare our proposed method, CcDPO, with existing multi-image DPO approaches across seven multi-image benchmarks. Our method consistently enhances the performance of both LLaVA-OV and Qwen2-VL.}
\resizebox{\textwidth}{!}{
\begin{tabular}{lccccccccc}
\toprule
\textbf{Models} & \textbf{Parameter} & \textbf{MuirBench} & \textbf{MIRB} & \textbf{BLINK} & \textbf{Mantis} & \textbf{NLVR2} & \textbf{MIBench} & \textbf{Q-Bench2} & \textbf{Average} \\
\midrule
GPT-4o~\cite{gpt-4v} & -& 62.3 & 53.0 & 60.1 & 62.7 & 88.8 & 71.8 & 74.5 & 67.6\\
\midrule
LLaVA-v1.5~\cite{li2024llava} & 7B & 19.9 & 28.4 & 37.1 & 41.9 & 52.1 & 40.9 & 53.9 & 39.2\\
Idefics2~\cite{laurenccon2024matters} & 8B & 26.1 & 33.0 & 45.2 & 48.9 & 86.9 & 29.7 & 57.0 & 46.6\\
Mantis-Idefics2~\cite{jiang2024mantis} & 8B & 44.5 & 41.8 & 49.1 & 57.1& 89.7 & 44.3 & 75.3 & 57.4\\
mPLUG-Owl3~\cite{ye2024mplug} & 8B & 39.6 & - & 50.3 & 63.1 & 90.8 & 54.5 & - & 59.6\\
Emu2-Chat~\cite{sun2024generative} & 37B & 33.6 & 27.2 & 36.2 & 37.8 & 58.2 & 39.7 & 65.3 & 42.6\\
InternVL2-8B~\cite{chen2024expanding} & 8B & 48.7 & 50.0 & 50.6 & 60.3 & 85.56 & 52.9 & - & 58.0\\
\midrule
LLaVA-OV~\cite{llava-ov} & 7B & 42.5 & 47.3 & 51.1 & 60.4 & 89.4 & 73.6 & 73.8 & 62.5\\ 
\quad + SFT & 7B & 45.4 & 48.9 & 53.4 & 64.9 & 89.0 & 71.9 & 75.7 & 64.1\\
\quad + MIA-DPO~\cite{mia-dpo} & 7B & 41.4 & 48.0 & 53.7 & 60.3 & 88.2 & 67.8 & 74.0 & 61.9\\
\rowcolor{cyan!10}
\quad + CcDPO (Ours) & 7B & \textbf{48.6} & \textbf{51.4} & \textbf{55.9} & \textbf{69.6} & \textbf{91.2} & \textbf{75.2} & \textbf{77.6}  & \textbf{67.1}\\
\quad $\Delta$ & - & \textcolor{customgreen}{\textbf{+6.1}} & \textcolor{customgreen}{\textbf{+4.1}} & \textcolor{customgreen}{\textbf{+4.8}} & \textcolor{customgreen}{\textbf{+9.2}} & \textcolor{customgreen}{\textbf{+1.8}} & \textcolor{customgreen}{\textbf{+1.6}} & \textcolor{customgreen}{\textbf{+3.8}} & \textcolor{customgreen}{\textbf{+4.6}}\\
\midrule
Qwen2-VL~\cite{qwen2vl} & 7B & 40.5 & 59.5 & 53.4 & 65.9 & 84.8 & 68.9 & 74.5 & 63.9\\
\quad + SFT & 7B & 43.1 & 59.8 & 54.7 & 64.9 & 85.2 & 69.4 & 74.1 & 64.5\\
\quad + MIA-DPO~\cite{mia-dpo} & 7B & 40.1 & 61.4 & 54.5 & 69.1 & 84.5 & 66.7 & 75.6 & 64.5\\
\rowcolor{cyan!10}
\quad + CcDPO (Ours) & 7B & \textbf{44.8} & \textbf{60.7} & \textbf{56.5} & \textbf{69.1} & \textbf{86.4} & \textbf{71.9} & \textbf{77.0} & \textbf{66.6}\\
\quad $\Delta$ & - & \textcolor{customgreen}{\textbf{+4.3}} & \textcolor{customgreen}{\textbf{+1.2}} & \textcolor{customgreen}{\textbf{+3.1}} & \textcolor{customgreen}{\textbf{+3.2}} & \textcolor{customgreen}{\textbf{+1.6}} & \textcolor{customgreen}{\textbf{+3.0}} & \textcolor{customgreen}{\textbf{+2.5}} & \textcolor{customgreen}{\textbf{+2.7}} \\
\bottomrule
\end{tabular}}
\vspace{-10pt}
\label{tab:multi_image_results}
\end{table*}

\begin{table*}[t]
\centering
\renewcommand{\arraystretch}{1.2}
\setlength{\tabcolsep}{3.5pt}
\caption{\textbf{Hallucination and preference alignment results.} We report metrics on our constructed multi-image AMBER benchmarks. Lower scores indicate better performance for CHAIR, Hal, and Cog, while higher is better for SCover. CcDPO achieves consistent improvements under both 4-image and 8-image settings, effectively reducing hallucinations in contextual multi-image understanding.}
\resizebox{0.95\linewidth}{!}{
\begin{tabular}{lccccc@{\hskip 6pt}cccc}
\toprule
\multicolumn{2}{c}{} & \multicolumn{4}{c}{\textbf{Context-AMBER (4 Images)}} & \multicolumn{4}{c}{\textbf{Context-AMBER (8 Images)}} \\
\cmidrule(lr){3-6} \cmidrule(lr){7-10}
\textbf{Models} & \textbf{Parameter} &
CHAIR$\downarrow$ & SCover$\uparrow$ & Hal$\downarrow$ & Cog$\downarrow$ &
CHAIR$\downarrow$ & SCover$\uparrow$ & Hal$\downarrow$ & Cog$\downarrow$ \\
\cmidrule(r){1-2} \cmidrule(lr){3-6} \cmidrule(l){7-10}
LLaVA-OV & 7B & 10.2 & 74.0\% & 31.8 & 2.6 & 50.6 & 10.3\% & 69.1 & 6.5 \\
\quad + MIA-DPO~\cite{mia-dpo} & 7B & 8.9 & 83.9\% & 29.8 & 2.1 & 28.2 & 36.7\% & 45.0 & 3.8 \\
\rowcolor{cyan!10}
\quad + CcDPO (Ours) & 7B & \textbf{3.7} & \textbf{100.0\%} & \textbf{15.3} & \textbf{1.2} & \textbf{15.3} & \textbf{83.3\%} & \textbf{27.5} & \textbf{2.1} \\
\quad $\Delta$ & - & \textcolor{customgreen}{\textbf{+6.5}} & \textcolor{customgreen}{\textbf{+26.0}} & \textcolor{customgreen}{\textbf{+16.5}} & \textcolor{customgreen}{\textbf{+1.4}} & \textcolor{customgreen}{\textbf{+35.3}} & \textcolor{customgreen}{\textbf{+70.0}} & \textcolor{customgreen}{\textbf{+41.6}} & \textcolor{customgreen}{\textbf{+4.4}} \\
\bottomrule
\end{tabular}
}
\vspace{-10pt}
\label{tab:hallucination}
\end{table*}

\begin{table}[t]
\centering
\renewcommand{\arraystretch}{1.1}
\setlength{\tabcolsep}{3.7pt}
\caption{\textbf{Performance on the image retrieval task from needle-in-a-haystack MM-NIAH~\cite{wang2024needle}.} We compare our proposed method, CcDPO, with DPO-based baselines across 1K–24K contexts, where the number of images ranges from a few to over a hundred. CcDPO consistently outperforms prior methods, demonstrating its strength in capturing fine-grained details in ultra-long image sequences.}
\begin{tabular}{lccccccccc}
\toprule
\textbf{Models} & \textbf{Parameter} & \textbf{1K} & \textbf{2K} & \textbf{4K} & \textbf{8K} & \textbf{12K} & \textbf{16K} & \textbf{24K} & \textbf{Average} \\
\midrule
LLaVA-OV~\cite{llava-ov}                     &  7B  & 89.2 & 88.1 & 82.3 & 71.2 & 65.0 & 60.9 & 45.0  & 71.7 \\
\quad + SFT                                  &  7B  & 92.0 & 93.9 & 88.2 & 80.7 & 74.8 & 69.4 & 49.9 & 78.4 \\
\quad + MIA-DPO~\cite{mia-dpo}               &  7B  & 93.9 & 94.6 & 90.5 & 85.1 & 75.5 & 68.8 & 59.4 & 81.1 \\
\rowcolor{cyan!10}
\quad + CcDPO (Ours)                &  7B  & \textbf{95.3} & \textbf{96.9} & \textbf{91.4} & \textbf{89.6}  & \textbf{78.8} & \textbf{74.5} & \textbf{64.7} & \textbf{84.5} \\
\quad $\Delta$ & - & \textcolor{customgreen}{\textbf{+6.3}} & \textcolor{customgreen}{\textbf{+8.8}} & \textcolor{customgreen}{\textbf{+9.1}} & \textcolor{customgreen}{\textbf{+18.4}} & \textcolor{customgreen}{\textbf{+13.8}} & \textcolor{customgreen}{\textbf{+13.6}} & \textcolor{customgreen}{\textbf{+19.7}} & \textcolor{customgreen}{\textbf{+12.8}} \\
\bottomrule
\end{tabular}
\label{tab:retrieval_image_needle}
\vspace{-5pt}
\end{table}

\begin{table*}[t]
\centering
\renewcommand{\arraystretch}{1.0}
\setlength{\tabcolsep}{4pt}
\caption{\textbf{Main results on single-image benchmarks.} We compare our CcDPO with existing DPO-based approaches across seven single-image benchmarks. Our CcDPO not only improves performance in multi-image settings but also preserves strong capabilities on single-image tasks.}
\label{tab:single_image_eval}
\resizebox{\linewidth}{!}{
\begin{tabular}{lccccccccccc}
\toprule
\textbf{Models} & \textbf{Parameter} & \textbf{MMStar} & \textbf{SQA} & \textbf{MMMU} & \textbf{POPE}  & \textbf{HallBench} & \textbf{MMB} & \textbf{OCR} & \textbf{AI2D} & \textbf{Avg.} \\
\midrule
LLaVA-v1.6~\cite{llava1.6}        & 7B  & 37.6 & 87.5 & 35.8 & 70.3 &51.6 & 69.8 & 53.7 & 67.0 & 59.1 \\
Qwen-VL-Chat~\cite{bai2023qwen}      & 7B  & 34.5 & 68.8 & 35.9 & 74.9 &39.2 & 61.8 & 48.8 & 63.0 & 53.3 \\
Idefics2~\cite{laurenccon2024matters}   & 8B  & 49.5 & 88.7 & 43.0 & 86.2 &- & 75.7 & - & 72.3 & 69.2 \\
OpenFlamingo~\cite{awadalla2023openflamingo}& 9B  & 36.9 & 44.8 & - & 52.6 &38.4 & 32.4 & 14.9 & 31.7 & 35.9 \\
InstructBLIP~\cite{instructblip}      & 13B & 32.7 & 54.1 & - & 86.1 &45.3 & 38.3 & 27.6 & 40.6 & 46.3 \\
Emu2-Chat~\cite{sun2024generative}         & 37B & 40.7 & 68.2 & 36.3 & 88.0 &- & 63.4 & 43.6 & 49.7 & 55.7 \\
\midrule
LLaVA-OV~\cite{llava-ov}               & 7B  & 58.7 &92.1 & \textbf{47.7} & 86.1 &52.9 & {81.8} & 47.3 & 81.6  & 68.5\\
\quad + SFT & 7B & 57.8 & 91.5 & 47.1 & \textbf{88.4} & 57.2 &  81.5 & 50.2 & 81.6 & 69.4\\
\quad + MIA-DPO~\cite{mia-dpo}           & 7B  & 57.4 &92.4 & 45.1 & {87.9} &55.4 & 80.9 & \textbf{52.1} & 81.5 & 69.0 \\
\rowcolor{cyan!10}
\quad + CcDPO (Ours)                   & 7B  & \textbf{59.5} &\textbf{92.6} & 45.7 & {86.6} &\textbf{58.4} & \textbf{81.9} & 51.0 & \textbf{82.1} & \textbf{69.7} \\
\midrule
Qwen2-VL~\cite{qwen2vl}          & 7B  & 57.8 & \textbf{84.1} & 50.6 & 85.9 &66.9 & 81.2 & \textbf{85.6} & 78.9 & 73.8 \\
\quad + SFT & 7B & 55.0 & 82.7 & 50.0 & 87.7 & 66.7 & 81.0 & 84.8 & 78.5 & 73.3\\
\quad + MIA-DPO~\cite{mia-dpo}                   & 7B  & 58.2 & {84.0} & 48.6 & \textbf{88.4} &62.7 & 80.8 & 85.1 & 78.9 & 73.3 \\
\rowcolor{cyan!10}
\quad + CcDPO (Ours)                   & 7B  & \textbf{58.7} & 82.8 & \textbf{50.7} & {87.1} &\textbf{68.8} & \textbf{81.6} & 83.5 & \textbf{79.7} & \textbf{74.1} \\
\bottomrule
\end{tabular}
}
\vspace{-10pt}
\end{table*}


\subsection{Main Results}
\vspace{-5pt}
\noindent \textbf{Results on General Multi-Image Tasks.}
As shown in Table~\ref{tab:multi_image_results}, we evaluate CcDPO across diverse multi-image benchmarks that span a wide range of reasoning skills. CcDPO consistently outperforms both the SFT baseline and other DPO-based methods on all datasets, with a notable gain of +4.8 points on BLINK, which focuses on multi-view and spatial reasoning. On the large-scale MuirBench dataset—where each sample contains an average of 4.3 images and up to 9 images—CcDPO achieves the largest improvement of +6.1 points, demonstrating its strength in modeling complex multi-image dependencies such as fine-grained perception, sequential cues, and holistic context. In contrast, MIA-DPO performs poorly on MuirBench, underscoring its limitations in capturing global context. The consistent gains across both LLaVA-OV and Qwen2-VL variants further validate the generality and effectiveness of our approach.

\noindent \textbf{Results on General Single-Image Tasks.}
While previous works~\cite{mia-dpo,jiang2024mantis} indicate that multi-image training can degrade single-image understanding, our CcDPO, in contrast, generally yields performance gains on most single-image datasets as shown in Table~\ref{tab:single_image_eval}, averaging +1.2 points for LLaVA-OV and +0.3 for Qwen2-VL. Visually-driven tasks like HallBench exhibit the largest improvements, up to +5.5 points under CcDPO. Conversely, for tasks with relatively low reliance on visual information, exemplified by ScienceQA~\cite{laurenccon2024matters}, our method showed no notable gains, and performance slightly declined.
These results highlight CcDPO's robustness: it not only excels in multi-image scenarios but also preserves, and often enhances, single-image capabilities. We attribute this success to our preference data design, which employs structured, per-image descriptions, thereby fostering precise understanding of individual images even within multi-image contexts.

\subsection{Ablation Studies}
\noindent \textbf{Comparing to SFT.} As shown in Table~\ref{tab:multi_image_results}, Table~\ref{tab:single_image_eval}, Table~\ref{tab:retrieval_image_needle}, CcDPO outperforms SFT across all benchmarks, achieving +3.0 on multi-image tasks, +0.6 on single-image tests, and a significant +6.1 gain on MM-NIAH. By integrating negative samples into DPO, CcDPO enhances discrimination between accurate and hallucinated outputs, improving fine-grained detail recognition and long-range dependency modeling while maintaining single-image performance. This demonstrates both robust generalization and superior contextual understanding, with negative sample integration proving essential to its performance advantages.

\noindent \textbf{Superior Context Scaling for Fine-Grained Detail Capture.}
As shown in Table~\ref{tab:retrieval_image_needle}, CcDPO significantly outperforms prior DPO-based methods and SFT baselines on the challenging MM-NIAH needle-in-a-haystack image retrieval task. Remarkably, CcDPO achieves consistent improvements across all tested context lengths (1K–24K), with its largest gain (+19.7 points) occurring at the maximum 24K context length compared to the baseline model. This demonstrates CcDPO's exceptional ability to: (1) Scale effectively to ultra-long image sequences, and (2) Capture fine-grained visual details critical for discriminating subtle differences in large image collections.
The +12.8-point average improvement underscores its robustness in handling large-scale multi-image contexts. The hierarchical preference optimization in CcDPO enables precise, context-aware understanding—an essential capability for processing extensive visual information.


\begin{table}[t]
\centering
\caption{\textbf{Ablation study of two-level CcDPO on MIBench and MIRB tasks} requiring perception, comparison, and reasoning across multiple images. Detailed task descriptions are in the Appendix~\ref{Benchmark}.}
\label{tab:MIBench}
\resizebox{\textwidth}{!}{
\begin{tabular}{l c c c c c c c c c c c c c c c}
\toprule
 & \multicolumn{8}{c}{\textbf{MIBench Benchmark}} & \multicolumn{4}{c}{\textbf{MIRB Benchmark}} \\
\cmidrule(lr){2-9} \cmidrule(lr){10-13}
\textbf{Models} & \textbf{GC} & \textbf{SD} & \textbf{TR} & \textbf{LR} & \textbf{FVR} & \textbf{TRI} & \textbf{VTK} & \textbf{TVK} & \textbf{Know.} & \textbf{Reas.} & \textbf{Perc.} & \textbf{M-Hop} \\
\midrule
LLaVA-OV-7B~\cite{llava-ov} & 87.7 & 85.9 & 72.6 & 74.5 & 96.5 & 77.5 & 42.7 & 67.1 & 70.0 & 44.0 & 50.0 & 12.0 \\
\quad + Context-Level        & 87.4 & 88.6 & 69.5 & 76.0 &  97.9 & 76.7 & 43.4 & 68.8 & 75.0 & 44.0 & 52.0 & 15.0 \\
\rowcolor{cyan!10}
\quad $\oplus$ Needle-Level         & 88.8 & 90.4 & 70.3 & 76.0 & 98.2 & 77.6 & 52.6 & 69.8 & 72.0 & 48.0 & 55.0 & 18.0 \\
\quad $\Delta$ 
& \textcolor{customgreen}{\textbf{+1.1}} 
& \textcolor{customgreen}{\textbf{+4.5}} 
& \textcolor{customgreen}{\textbf{-2.3}} 
& \textcolor{customgreen}{\textbf{+2.5}} 
& \textcolor{customgreen}{\textbf{+0.1}} 
& \textcolor{customgreen}{\textbf{+1.7}} 
& \textcolor{customgreen}{\textbf{+9.9}}
& \textcolor{customgreen}{\textbf{+2.7}} 
& \textcolor{customgreen}{\textbf{+2.0}} 
& \textcolor{customgreen}{\textbf{+4.0}} 
& \textcolor{customgreen}{\textbf{+5.0}} 
& \textcolor{customgreen}{\textbf{+6.0}}  \\
\midrule
\quad + SFT                 & 87.9 & 88.2 & 71.0 & 75.5  & 87.8 & 77.4 & 38.5 & 68.3 & 75.0 & 43.0 & 53.0 & 9.0 \\
\quad + MIA-DPO~\cite{mia-dpo} & 85.8 & 87.2 & 63.4 & 67.9 & 94.9 & 67.6 & 42.0 & 59.6 & 73.0 & 50.0 & 44.0 & 11.0 \\
\bottomrule
\end{tabular}}
\vspace{-20pt}
\end{table}

\begin{table}[t]
\centering
\caption{\textbf{Ablation study of CcDPO on the MuirBench dataset} across all sub-datasets, demonstrating significant performance gains on most subsets. The symbol $\oplus$ stands for method superposition.}
\label{tab:MuirBench}
\resizebox{\textwidth}{!}{
\begin{tabular}{l c *{12}{c}}
\toprule
\textbf{Models} & \textbf{Overall} 
  & \rotatebox{90}{{Action}} 
  & \rotatebox{90}{{Similarity}} 
  & \rotatebox{90}{{Cartoon}}
  & \rotatebox{90}{{Counting}} 
  & \rotatebox{90}{{Diagram}} 
  & \rotatebox{90}{{Difference}} 
  & \rotatebox{90}{{Geographic}} 
  & \rotatebox{90}{{I-T Match}} 
  & \rotatebox{90}{{Ordering}} 
  & \rotatebox{90}{{Scene}} 
  & \rotatebox{90}{{Grounding}} 
  & \rotatebox{90}{{Retrieval}} \\
\midrule
LLaVA-OV-7B~\cite{llava-ov} & 42.5 & 35.9 & 33.1 & 35.8 & 24.7 & 55.0 & 30.0 & 46.0 & 46.9 & 20.3 & 72.0 & 29.7 & 47.6 \\
\quad + Context-Levl & 44.8 & 36.6 & 30.3 & 37.1 & 35.5 & 57.0 & 37.9 & 41.0 & 48.3 & 15.6 & 65.1 & 28.3 & 48.3 \\
\quad $\oplus$ Needle-Levl-TDPO & 47.8 & 42.3 & 48.0 & 34.6 & 38.0 & 58.3 & 40.3 & 41.0 & 55.4 & 14.1 & 68.9 & 30.1 & 45.9 \\
\rowcolor{cyan!10}
\quad $\oplus$ Needle-Levl-VDPO & 48.6 & 44.5 & 49.5 & 33.3 & 39.3 & 60.1 & 39.1 & 41.0 & 56.7 & 13.0 & 70.4 & 28.6 & 46.6 \\
\quad $\Delta$ & \textcolor{customgreen}{\textbf{+6.1}} & \textcolor{customgreen}{\textbf{+8.6}} & \textcolor{customgreen}{\textbf{+16.4}} & \textcolor{customgreen}{\textbf{-2.5}} & \textcolor{customgreen}{\textbf{+14.6}} & \textcolor{customgreen}{\textbf{+5.1}} & \textcolor{customgreen}{\textbf{+9.1}} & \textcolor{customgreen}{\textbf{-5.0}} & \textcolor{customgreen}{\textbf{+9.8}} & \textcolor{customgreen}{\textbf{-7.3}} & \textcolor{customgreen}{\textbf{-1.6}} & \textcolor{customgreen}{\textbf{-1.1}} & \textcolor{customgreen}{\textbf{-1.0}}  \\
\midrule
Qwen2-VL-7B~\cite{qwen2vl} & 40.5 & 40.8 & 46.4 & 41.0 & 39.7 & 41.9 & 34.4 & 25.0 & 54.3 & 9.3 & 65.0 & 28.5 & 20.5 \\
\quad + Context-Levl & 42.1 & 41.0 & 45.0 & 42.3 & 39.8 & 46.2 & 37.9 & 22.0 & 56.9 & 14.1 & 65.1 & 27.4 & 21.0 \\
\quad $\oplus$ Needle-Levl-TDPO & 42.3 & 39.6 & 43.9 & 41.0 & 39.7 & 46.5 & 39.1 & 22.0 & 56.3 & 15.7 & 66.7 & 26.2 & 22.9 \\
\rowcolor{cyan!10}
\quad $\oplus$ Needle-Levl-VDPO & 44.8 & 43.9 & 46.9 & 38.5 & 39.9 & 53.8 & 37.9 & 20.0 & 61.4 & 18.8 & 69.9 & 23.8 & 22.9 \\
\quad $\Delta$ & \textcolor{customgreen}{\textbf{+4.3}} & \textcolor{customgreen}{\textbf{+3.1}} & \textcolor{customgreen}{\textbf{+0.5}} & \textcolor{customgreen}{\textbf{-2.5}} & \textcolor{customgreen}{\textbf{+0.2}} & \textcolor{customgreen}{\textbf{+11.9}} & \textcolor{customgreen}{\textbf{+3.5}} & \textcolor{customgreen}{\textbf{-3.0}} & \textcolor{customgreen}{\textbf{+7.1}} & \textcolor{customgreen}{\textbf{+4.9}} & \textcolor{customgreen}{\textbf{+5.4}} & \textcolor{customgreen}{\textbf{-4.7}} & \textcolor{customgreen}{\textbf{+2.4}} \\
\bottomrule
\end{tabular}}
\vspace{-15pt}
\end{table}


\noindent \textbf{Effectiveness of Context-Level Optimization.} The Context-Level DPO  enhances alignment between language responses and the holistic visual context across image sequences. As shown in Table~\ref{tab:MuirBench}, this yields significant improvements on sub-tasks requiring global reasoning, including: Diagram Understanding (+5.1), Image-Text Matching (+9.8), Similarity Matching (+16.4). These gains reflect the module's ability to: (1) Capture semantic relationships across images, and (2) Maintain coherent multi-image descriptions through consistent attribute attribution. Notably, Scene Understanding and Retrieval tasks also benefit from improved global alignment, confirming that such optimization effectively reduces context omission and conflation errors in complex visual sequences.

\noindent \textbf{Effectiveness of Needle-Level Optimization.}
The Needle-Level DPO improves the model’s ability to capture fine-grained visual cues by contrasting localized content. As shown in Table~\ref{tab:MIBench} and Table~\ref{tab:MuirBench}, this is especially effective for tasks requiring detailed comparisons across images. In MIBench, the VTK task—where the model must link information across image cells—shows a large gain of +9.9, demonstrating that our visual preference signals help focus on factual visual details. Similarly, in MuirBench, we observe strong gains on Action Understanding (+8.6), Counting (+14.6), and Difference Spotting (+9.1), all of which depend on localized perception. These results indicate that Needle-Level DPO significantly boosts the model’s perceptual grounding and resistance to detail-level hallucinations, complementing the context-level dpo for better multi-image understanding.

\noindent \textbf{Limitations}
While CcDPO is primarily designed for general multi-image reasoning, it does not explicitly model temporal dependencies, which may limit its performance on video-like inputs. However, the framework is readily extensible to such data by incorporating temporal-aware prompts. Similarly, although OCR supervision is limited in our current dataset, CcDPO can be naturally enhanced with targeted text-centric preference data in future work.

\section{Conclusion}
This work introduces \textbf{CcDPO}, a two-level preference optimization method for enhancing multi-image understanding in MLLMs. By decoupling learning into context-level and needle-level stages, CcDPO addresses key hallucination issues including context omission, conflation, and detail misinterpretation. The context-level module promotes holistic sequence comprehension via structured caption preferences, while the needle-level module strengthens fine-grained perception through visual prompts and contrastive supervision. To support optimization, we construct \textbf{MultiScope-42k}, a large-scale dataset with automatically generated multi-level preference pairs. Experiments across seven multi-image benchmarks show that CcDPO achieves consistent improvements over SFT and prior DPO variants, confirming its effectiveness in aligning MLLMs with both global and local visual semantics.

\medskip
{
\bibliographystyle{unsrt}
\bibliography{neurips_2025.bib}
}
\newpage
\appendix
\section*{\centerline{Technical Appendices}}
\vspace{-10pt}
In this appendix, we provide additional materials to support a more comprehensive understanding of our proposed method and dataset.
\textbf{In Sec.~A}, we detail the low-cost construction pipeline of \textbf{MultiScope-42k} and conduct comparative data analysis with MIA-DPO, including token length distributions, word cloud statistics, and supervision source breakdown. We also clarify the image source overlap between training data and benchmarks to ensure fair evaluation.
\textbf{In Sec.~B}, we summarize all benchmarks used in evaluation, including seven multi-image and eight single-image benchmarks.
\textbf{In Sec.~C}, we provide additional experimental results, including ablation studies on training data volume and supervision granularity to assess their impact on model performance.
\textbf{In Sec.~D}, we present qualitative observations and visualizations of preference pairs.

\section{MultiScope-42k: A Context-to-Cue Captioning DPO Dataset}
\subsection{Low-cost Question-Answer Pair Construction}
Constructing high-quality instruction-response preference pairs for multi-image learning traditionally requires extensive manual annotation, especially when capturing subtle context dynamics or region-level semantics. To address this bottleneck, we design a low-cost, fully automated pipeline for question-answer pair construction, enabling efficient and scalable data generation with broad coverage and controlled distributional properties.

\noindent\textbf{Automated Caption Pool Sampling.}
We first leverage existing vision-language datasets—LLaVA-23K~\cite{liu2023visual}, MDVP~\cite{lin2024draw}, and MVC~\cite{wu2025symmetrical}—to construct a diverse caption pool, containing both image-level and region-level descriptions. By decoupling question construction from caption generation, we are able to sample visual contexts and their aligned captions independently, facilitating large-scale composition of input-output examples.

\noindent\textbf{Structured QA Formatting.}
Given a sampled image or image sequence, we construct templated instructions (e.g., “Describe the following images” or “Please describe the marked area in each image”) to form queries. For the corresponding answers, we use structured formats that encourage compositional reasoning and grounding, such as:
\begin{adjustwidth}{0em}{0pt}
\centering
\texttt{[ For Image 1: <caption 1>, For Image 2: <caption 2>, ...]}
\end{adjustwidth}
\begin{adjustwidth}{0em}{0pt}
\centering
\texttt{[ For the marked area of Image X: <caption X>, ...]}
\end{adjustwidth}
This approach allows flexible variation in image number, visual scope, and response granularity—supporting both context- and region-level supervision.

\noindent\textbf{Controlled Perturbation for Preference Learning.}
To generate preference pairs without additional labeling, we apply lightweight perturbation strategies to the answer side only. Specifically:\\
\noindent \textbullet\ \textit{(1) Truncation and swapping:} simulate omissions and misalignments in context-level answers.\\
\noindent \textbullet\ \textit{(2) Region mismatches:} in needle-level samples simulate detail hallucination.\\
These perturbations require~\textbf{no human involvement} yet introduce controlled errors mirroring real-world MLLM failure modes, enabling scalable preference pair generation with low cost.

\noindent\textbf{Efficient Coverage of Diverse Visual Distributions.}
Our automated sampling supports stratified control over image domains, scene compositions, and region attributes. This results in the MultiScope-42k dataset, a large-scale, distribution-aware corpus covering a wide variety of multi-image tasks. Its diversity in visual layout and semantic granularity ensures robust preference supervision across image types and reasoning levels.

\noindent Overall, our pair construction strategy eliminates the need for dense manual annotation while producing rich and challenging preference data at scale—offering a practical solution for instruction tuning in multi-image multimodal models.

\subsection{Data Analysis}
\begin{table}[t!]
\centering
\caption{Summary statistics of MultiScope-42k by supervision level and image source, including total pairs, number of images per instance, and average token lengths of chosen and rejected responses.}
\resizebox{\textwidth}{!}{
\begin{tabular}{l l c c c c}
\toprule
\textbf{Level} & \textbf{Image Source} & \textbf{Total} & \textbf{Images Range} & \textbf{Avg. Chosen Len.} & \textbf{Avg. Rejected Len.} \\
\midrule
Context-Level & COCO-2014   & 27.3k & [2, 5] & 285.24 & 165.45 \\
Needle-Level-TDPO  & COCO-2017   & 10.8k & [2, 4] & 173.35 & 173.97 \\
Needle-Level-VDPO  & Flickr30k   & 3.7k  & [2, 4] & 66.09  & 65.98  \\
\bottomrule
\end{tabular}}
\label{tab:token-length}
\end{table}

To better understand the characteristics of our preference supervision, we conduct a comparative analysis of MultiScope-42k and MIA-DPO from both lexical and structural perspectives.

\begin{figure*}[t!]
  \centering
    \includegraphics[width=1\textwidth]{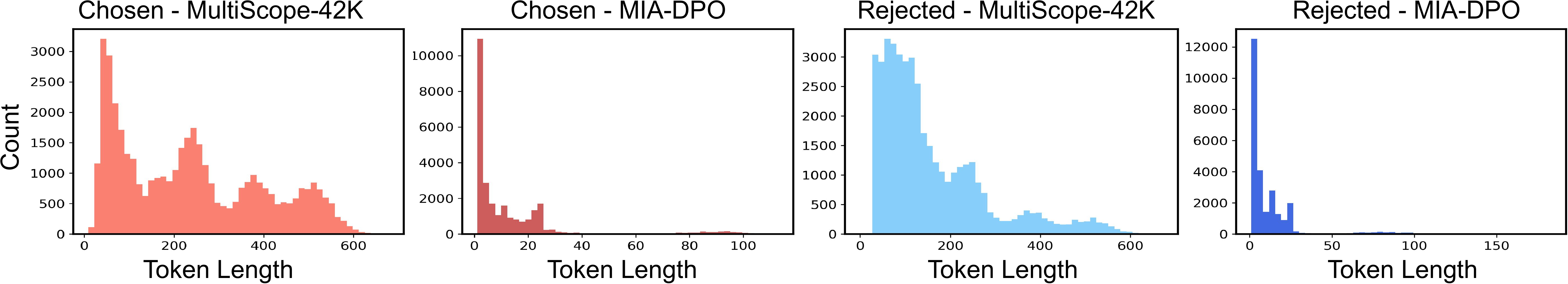}
  \caption{
  {Token length distributions of chosen and rejected responses in our MultiScope-42k and MIA-DPO~\cite{mia-dpo}. MultiScope-42k exhibits significantly longer and more diverse answers, while MIA-DPO responses remain short and concentrated, indicating a simpler response pattern.}
  }
  \label{fig:tokenlength}
\end{figure*}

\noindent \textbf{Token Length Distributions.}
Fig.~\ref{fig:tokenlength} presents the token length histograms of both \textit{chosen} and \textit{rejected} responses. MultiScope-42k responses are significantly longer on average and display a wider spread, with many responses exceeding 400 tokens. This reflects the dataset's multi-stage, image-wise captioning format and compositional design. In contrast, MIA-DPO responses are short and concentrated, with the majority under 30 tokens. This further suggests that MultiScope-42k provides richer and more diverse supervision signals, especially for multi-image reasoning.

\begin{figure*}[t!]
  \centering
    \includegraphics[width=1\textwidth]{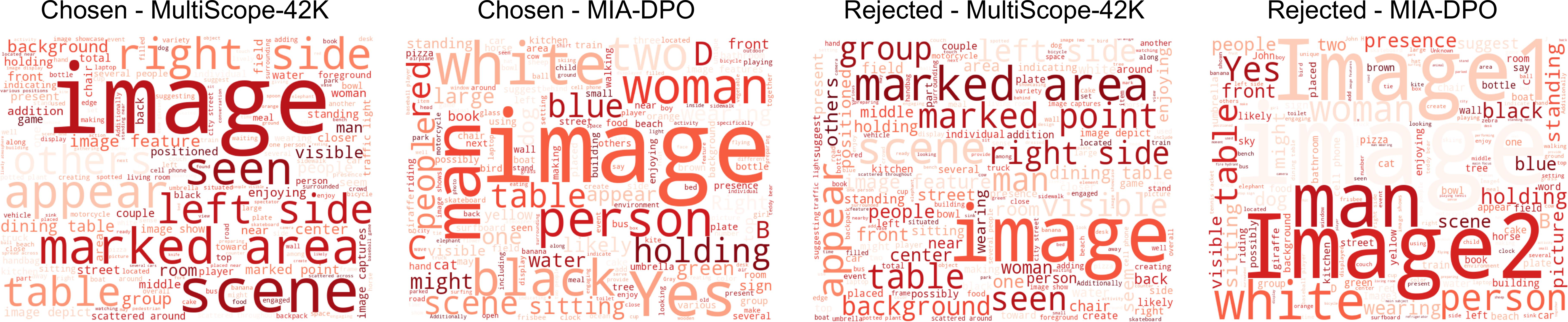}
  \caption{The dataset word cloud comparison between our MultiScope-42k and MIA-DPO~\cite{mia-dpo}.
  }
  \label{fig:wordcloud}
\end{figure*}

\noindent \textbf{Linguistic Focus via Word Cloud.}
We visualize the answer sets of both datasets using word clouds. As shown in Fig.~\ref{fig:wordcloud}, MultiScope-42k responses prominently feature structured and spatially grounded expressions such as \textit{“image,” “marked area,” “left,” “right,” “foreground,” “scene”}, indicating a strong alignment with multi-image, region-specific prompts. In contrast, MIA-DPO emphasizes atomic visual concepts (e.g., \textit{“man,” “table,” “dog,” “red”}), which are well-suited for single-image tasks but lack explicit inter-image reference or structural composition.


\subsection{Dataset Independence Statement}
To ensure fair and unbiased evaluation, we verify that the benchmarks used for testing do not overlap with the data sources involved in model training.
We employ three types of annotated data for two-level dpo training. 
\noindent {(1) Image-level captions} are sourced from \textbf{LLaVA-23K}~\cite{liu2023visual}, which is derived from the COCO-2014 dataset. 
\noindent {(2) Region-level captions} come from the \textbf{MDVP} dataset~\cite{lin2024draw}, based on COCO-2017. 
\noindent {(3) Visual contrastive preference pairs} are taken from \textbf{MVC}~\cite{wu2025symmetrical}, whose image sources include \textbf{CounterCurate}~\cite{zhang2024countercurate} and \textbf{FineCops-Ref}~\cite{liu2024finecops}.

We conducted a thorough review of all evaluation benchmarks for potential overlap with these training sources (COCO-2014, COCO-2017, Flickr30k). While most multi-image benchmarks appear to be independently constructed, we conservatively flag the following for partial or uncertain overlap:

\noindent \textbf{MIRB}~\cite{mirb} explicitly uses COCO images (e.g., for counting), and includes content from ImageNet-R, OpenFoodFact, Bitton et al., and arXiv. 
\noindent \textbf{MIBench}~\cite{liu2024mibench} comprises multiple public datasets, including VrR-VG, which inherits COCO images via Visual Genome. 
\noindent \textbf{MUIRBench}~\cite{wang2024muirbench} integrates existing (e.g., SeedBench, IconQA), derived (e.g., NLVR2, MMBench), and newly collected datasets. While its new data is COCO-free, MMBench and IconQA are known to include COCO images.

A summary of benchmark image sources is provided below:

\noindent \textbf{MUIRBench} combines new and repurposed datasets; some derived components include COCO. 
\noindent \textbf{MIRB} includes COCO, ImageNet-R, and other web-sourced content. 
\noindent \textbf{BLINK} uses synthetic and controlled real-world images. 
\noindent \textbf{Mantis-Eval} draws from web queries and manual composition, with no COCO usage. 
\noindent \textbf{NLVR2} uses Flickr images and is COCO-independent. 
\noindent \textbf{Q-Bench2} is based on IQA datasets like KonIQ-10k and BID, unrelated to COCO or Flickr30k. 
\noindent \textbf{MIBench} includes VrR-VG, which partially overlaps with COCO.

Although a few benchmarks partially overlap with COCO or Flickr30k, we argue that this does not compromise evaluation fairness for two reasons. First, our main baseline \textbf{MIA-DPO} is also trained on \textbf{LLaVA-23K}, which uses COCO images, placing all methods on comparable footing. Second, our DPO training is designed for \textbf{image-caption preference alignment} rather than question answering. Since our model is never exposed to multi-image QA or instruction tasks during training, performance gains cannot be attributed to direct data memorization.
Together, these considerations support the integrity and fairness of our evaluation protocol, even in the presence of partial dataset overlap.

\section{Benchmark Sources}\label{Benchmark}
\subsection{More Details on the Construction of Context-AMBER-1K}
We design a caption generation task as a proxy to systematically evaluate MLLMs' multi-image understanding, identifying three fundamental hallucination types—\textit{Context Omission}, \textit{Context Conflation}, and \textit{Detail Misinterpretation}—that critically degrade model performance.
To enable this evaluation, we construct \textbf{Context-AMBER-1K} by systematically concatenating images from the single-image AMBER dataset~\cite{amber} into sequences of two types: short-context sequences with 4 images and long-context sequences with 8 images. Each input is paired with the prompt: \texttt{``Please sequentially describe each of the images shown above. Use the format: For Image *:<description>.''}.

To evaluate \textbf{Detail Misinterpretation}, we employ three established metrics: (a) CHAIR~\cite{rohrbach2018object}; (b) Response-level hallucination rate (Hal); and (c) Cognition-based hallucination (Cog). Each generated caption is matched to its corresponding image using regular expressions to extract per-image descriptions. We then compute hallucination scores for each image separately and average these scores across the entire image sequence. To assess \textbf{Context Omission} and \textbf{Context Conflation}, we introduce two rule-based penalization strategies:

(1) For context omission, if fewer than \( N \) captions are generated (e.g., the response ends prematurely at \texttt{\{For Image N-1: <caption N-1>\}}), we explicitly pad the output to \texttt{\{For Image N: <caption N-1>\}}) to maintain a consistent structure. Missing or incomplete captions are heavily penalized. Additionally, we introduce the (d) Sequence Coverage Rate (SCover), a metric that evaluates caption completeness across image sequences and quantifies the degree of context omission.

(2) For context conflation, we include the keyword \texttt{``sequentially''} in the prompt to encourage models to describe images in order. If the model cannot clearly associate each description with its corresponding image (e.g., \texttt{For Image 1:<caption1>, For Image 3:<caption2>, For Image 2:<caption3>}), the resulting hallucination scores will be higher for out-of-order descriptions compared to their correct GTs, as the mismatch leads to a larger discrepancy.

\subsection{Multi-Image Benchmarks}
We employ seven multi-image benchmarks—{MUIRBench}~\cite{wang2024muirbench}, {MIRB}~\cite{mirb}, BLINK~\cite{blink}, Mantis-Eval~\cite{jiang2024mantis}, NLVR2~\cite{nlvr2}, {Q-Bench2}~\cite{q-bench2}, and MIBench~\cite{liu2024mibench}—to holistically evaluate multi-image reasoning across four key dimensions: co-reference alignment, fine-grained comparison, contextual reasoning, and temporal understanding.

\noindent \textbf{MUIRBench}~\cite{wang2024muirbench}  is a comprehensive benchmark specifically designed to evaluate the robustness of multimodal large language models (MLLMs) in multi-image understanding scenarios. It comprises 2,600 multiple-choice questions and 11,264 images, averaging 4.3 images per instance. The benchmark covers 12 distinct multi-image understanding tasks—including action understanding, diagram reasoning, geographic comprehension, and visual retrieval—spanning 10 diverse multi-image relation types such as temporal, narrative, and scene-multiview relations.
To ensure both comprehensiveness and robustness, MUIRBench adopts a pairwise design: each standard (answerable) question is paired with an unanswerable variant with minimal semantic perturbations. This enables fine-grained assessments of both reasoning capability and abstention behavior.

\noindent \textbf{MIRB}~\cite{mirb} is a comprehensive benchmark designed to evaluate vision-language models (VLMs) on four distinct aspects of multi-image understanding: perception, visual world knowledge, reasoning, and multi-hop reasoning. It comprises 925 multi-image questions across these categories, averaging 3.78 images per question, with some tasks requiring up to 42 images for complex reasoning. Unlike prior benchmarks that reuse video frames, MIRB independently sources images from real-world domains, such as code snippets, sightseeing scenes, food ingredient lists, and arXiv papers, ensuring diverse and challenging visual contexts.
The benchmark includes a wide array of tasks: image jigsaw reconstruction, object counting, attribute matching \textbf{(Perception)}; food label comparison and geographic recognition \textbf{(Knowledge)}; visual analogy, code understanding, 3D scene analysis \textbf{(Reasoning)}; and synthetic logic chains and citation lookups \textbf{(Multi-Hop)}. Each question is formulated to necessitate reasoning across multiple images rather than from a single image.

\noindent \textbf{MIBench}~\cite{liu2024mibench} is a large-scale benchmark designed to comprehensively evaluate the fine-grained multi-image understanding abilities of multimodal large language models (MLLMs). It categorizes multi-image inputs into three representative scenarios—Multi-Image Instruction (MII), Multimodal Knowledge-Seeking (MKS), and Multimodal In-Context Learning (MIC)—covering a total of 13 distinct tasks and 13,000 annotated samples.
In the MII setting, the model must perform perception, comparison, and reasoning over multiple images across five tasks: general comparison \textbf{(GC)}, subtle difference \textbf{(SD)}, visual referring \textbf{(VR)}, temporal reasoning \textbf{(TR)}, and logical reasoning \textbf{(LR)}. The MKS scenario evaluates the model’s ability to extract and align information from interleaved image-text knowledge sources through four tasks: fine-grained visual recognition \textbf{(FVR)}, text-rich image VQA \textbf{(TRI)}, vision-linked textual knowledge \textbf{(VTK)}, and text-linked visual knowledge \textbf{(TVK)}. Finally, the MIC setting assesses multimodal in-context learning across four tasks, including close-ended and open-ended VQA, hallucination mitigation, and demo-based task learning.

\noindent \textbf{BLINK}~\cite{blink} tests rapid visual cognition through perceptual similarity, forensic analysis, and spatiotemporal matching. It includes tightly-controlled multi-image tasks such as depth estimation, object matching, and outlier detection, with an emphasis on speed and perceptual accuracy.

\noindent \textbf{Mantis-Eval}~\cite{jiang2024mantis} introduces 217 multi-image tasks curated for conceptual inference, including abstract reasoning over physical quantities such as number, size, and weight. It combines both multiple-choice and open-ended questions, drawing from web-sourced image sets manually organized into logical visual groupings.

\noindent \textbf{NLVR2}~\cite{nlvr2} (Natural Language Visual Reasoning) assesses a model’s ability to verify textual hypotheses against a pair of images. Each sample requires binary classification (True/False) over whether the provided statement is consistent with both images, making it a canonical test for visual entailment and compositional reasoning.

\noindent \textbf{Q-Bench2}~\cite{q-bench2} is a diagnostic benchmark tailored for evaluating visual quality perception and comparative assessment across image sets. It challenges models to identify subtle visual artifacts, distortions, or improvements between similar images. Our evaluation is based on the Q-Bench2-A1-dev subset, which emphasizes multi-image multiple-choice assessments for perceptual judgment.
\begin{table}[t!]
\centering
\caption{\textbf{Benchmark Sources.} We have included detailed information for all the multi-image and single-image benchmarks tested in the paper in the table.}
\label{tab:benchmarks}
\resizebox{\linewidth}{!}{
\begin{tabular}{llccc}
\toprule
\textbf{Setting} & \textbf{Models} & \textbf{Evaluation Metric} & \textbf{Number} & \textbf{Source} \\
\midrule
\multirow{5}{*}{\textbf{Multi-Image Benchmark}} 
& MUIRBench~\cite{wang2024muirbench} & Multiple Choice & 2,600 & \textcolor{magenta}{\textbf{MUIRBench}} \\
& MIRB~\cite{mirb} & Multiple Choice & 925 & \textcolor{magenta}{\textbf{MIRB}} \\
& MIBench~\cite{liu2024mibench} & Multiple Choice & 13,000 & \textcolor{magenta}{\textbf{MIBench}} \\
& BLINK~\cite{blink} & Multiple Choice & 3,807 & \textcolor{magenta}{\textbf{BLINK}} \\
& NLVR2~\cite{nlvr2} & Multiple Choice & 6,967 & \textcolor{magenta}{\textbf{NLVR2}} \\
& Q-Bench2~\cite{q-bench2} & Multiple Choice & 1,000 & \textcolor{magenta}{\textbf{Q-Bench2}} \\
& Mantis-Eval~\cite{jiang2024mantis} & Multiple Choice & 217 & \textcolor{magenta}{\textbf{Mantis-Eval}} \\
\midrule
\multirow{7}{*}{\textbf{Single-Image Benchmark}} 
& MMStar~\cite{mmstar} & Multiple Choice & 1,500 & \textcolor{magenta}{\textbf{MMStar}} \\
& MMMU~\cite{yue2024mmmu} & Multiple Choice & 1,050 & \textcolor{magenta}{\textbf{MMMU}} \\
& Sci-QA~\cite{sqa} & Multiple Choice & 4,241 & \textcolor{magenta}{\textbf{ScienceQA}} \\
& POPE~\cite{pope} & Yes/No & 9,000 & \textcolor{magenta}{\textbf{POPE}} \\
& HallBench~\cite{HallBench} & Yes/No & 951 & \textcolor{magenta}{\textbf{HallusionBench}} \\
& MMB~\cite{liu2024mmbench} & Multiple Choice & 1,164 & \textcolor{magenta}{\textbf{MMBench}} \\
& OCR~\cite{liu2024ocrbench} & VQA & 1,000 & \textcolor{magenta}{\textbf{OCRBench}} \\
& AI2D~\cite{ai2d} & Multiple Choice & 3,090 & \textcolor{magenta}{\textbf{AI2D}} \\
\bottomrule
\end{tabular}}
\vspace{-10pt}
\end{table}
\vspace{-20pt}
\subsection{Single-Image Benchmarks}
We test the model on eight representative single-image benchmarks assess specific multimodal capabilities: (1) Academic/Scientific Reasoning: {MMMU}~\cite{yue2024mmmu}, {MMStar}~\cite{mmstar}, {ScienceQA}~\cite{sqa}, (2) Diagram Understanding: {AI2D}~\cite{ai2d}, (3) Robustness against hallucinations: {POPE}~\cite{pope}, {HallBench}~\cite{HallBench}, (4) General Multimodal Abilities: {MMBench}~\cite{liu2024mmbench}, (5) Text Recognition: {OCRBench}~\cite{liu2024ocrbench}. The results on this diverse set of benchmarks demonstrate the effectiveness of the proposed method, particularly in multi-image settings, confirming significant performance improvements.

\noindent \textbf{MMMU}~\cite{yue2024mmmu} (Massive Multimodal Multitask Understanding) includes over 10k university-level questions from 30+ disciplines such as physics, medicine, and art. It requires detailed reasoning over image-text inputs and is designed to evaluate advanced academic-level understanding.

\noindent \textbf{MMStar}~\cite{mmstar} is a comprehensive diagnostic benchmark covering various sub-tasks such as OCR, VQA, and caption grounding, offering structured and hierarchical annotations across domains like natural science, medicine, and design.

\noindent \textbf{ScienceQA}~\cite{sqa} contains over 21k science questions aligned with elementary and middle school curricula, involving images such as diagrams and charts. It tests the model’s capability to perform science-related visual reasoning in a multimodal format.

\noindent \textbf{AI2D}~\cite{ai2d} (Allen Institute Diagram) features manually annotated science diagrams with associated multiple-choice questions. It focuses on assessing the model's understanding of labeled structures and their functional roles within the image.

\noindent \textbf{POPE}~\cite{pope} (Position and Object-level Prompt Evaluation) is designed to test a model’s resistance to hallucinations. It uses minimally perturbed prompts to identify failure cases in positional grounding and object identification, highlighting model robustness.

\noindent \textbf{HallBench}~\cite{HallBench} provides a structured framework to measure hallucination frequency and grounding quality by comparing model outputs with annotated ground truths. It supports fine-grained scoring across categories such as incorrect object mentions or unsupported claims.

\noindent \textbf{MMBench}~\cite{liu2024mmbench} is a general-purpose evaluation benchmark comprising questions across 11 modalities including VQA, captioning, OCR, and commonsense reasoning. It uses GPT-4-based grading to ensure high-fidelity evaluation of answer correctness.

\noindent \textbf{OCRBench}~\cite{liu2024ocrbench} specifically targets the model's capability to recognize and reason about text in the visual domain, covering a range of document layouts, fonts, and multilingual content with both exact-match and reasoning-based questions.

\section{More Experiments}
\noindent \textbf{Data Scale Alignment with MIA-DPO.}
To assess the impact of training size and ensure a fair comparison with MIA-DPO, we conduct an ablation in Tab.~\ref{tab:multi_image_results_datasize} using a similar total number of preference pairs. Specifically, we randomly sample 13.6k from our 27.3k Context-Level pairs and combine them with the fixed 14.5k Needle-Level data, resulting in a 28.1k training set—comparable to MIA-DPO’s 28.9k.
Notably, under this matched training data size, our \textbf{CcDPO} still outperforms MIA-DPO across all benchmarks, demonstrating the effectiveness of our structured, dual-level supervision. In particular, the reduced-context setting yields better performance on BLINK and Q-Bench2, suggesting that a relatively higher proportion of needle-level data may benefit fine-grained perceptual tasks. On the other hand, performance on context-heavy benchmarks drops slightly, likely due to weaker global context modeling.
Overall, training with the full 42k preference set (28.1k Context-Level + 14.5k Needle-Level) leads to the best average performance. These results highlight the advantage of high-quality, large-scale supervision, while also revealing a trade-off between contextual alignment and perceptual precision.

\begin{table*}[t]
\centering
\renewcommand{\arraystretch}{1.0}
\setlength{\tabcolsep}{3.5pt}
\caption{\textbf{Ablation on training data volume.} 
To match MIA-DPO's training data size, we downsample our Context-Level data to 13.6k while keeping needle-level data fixed. The results reveal a trade-off between modeling global context and capturing fine-grained details: while reduced Context-Level data leads to performance drops on most multi-image tasks, benchmarks like BLINK and Q-Bench2—focused on localized perception—benefit from a higher proportion of needle-level data.
}
\resizebox{\textwidth}{!}{
\begin{tabular}{lccccccccc}
\toprule
\textbf{Models} & \textbf{Data Size} & \textbf{MuirBench} & \textbf{MIRB} & \textbf{BLINK} & \textbf{Mantis} & \textbf{NLVR2} & \textbf{MIBench} & \textbf{Q-Bench2} & \textbf{Average} \\
\midrule
LLaVA-OV~\cite{llava-ov} & - & 42.5 & 47.3 & 51.1 & 60.4 & 89.4 & 73.6 & 73.8 & 62.5\\ 
\quad + MIA-DPO~\cite{mia-dpo} & 28.9K & 41.4 & 48.0 & 53.7 & 60.3 & 88.2 & 67.8 & 74.0 & 61.9\\
\quad + CcDPO (Ours) & 28.1K & 46.7  &51.2  & \textbf{56.5}  & 69.1  & 90.7 &72.1  & \textbf{79.3}  & 66.5  \\
\rowcolor{cyan!10}
\quad + CcDPO (Ours) & 41.8K & \textbf{48.6} & \textbf{51.4} & {55.9} & \textbf{69.6} & \textbf{91.2} & \textbf{75.2} & {77.6}  & \textbf{67.1}\\
\quad $\Delta$ & - & \textcolor{customgreen}{\textbf{+6.1}} & \textcolor{customgreen}{\textbf{+4.1}} & \textcolor{customgreen}{\textbf{+4.8}} & \textcolor{customgreen}{\textbf{+9.2}} & \textcolor{customgreen}{\textbf{+1.8}} & \textcolor{customgreen}{\textbf{+1.6}} & \textcolor{customgreen}{\textbf{+3.8}} & \textcolor{customgreen}{\textbf{+4.6}}\\
\bottomrule
\end{tabular}}
\vspace{-10pt}
\label{tab:multi_image_results_datasize}
\end{table*}

\noindent \textbf{Ablation of Training Strategies.}  
We investigate whether our multi-stage training strategy—first training on Context-Level data, followed by Needle-Level supervision—is more effective than a single-stage approach that mixes both types of data from the beginning. As shown in Tab.~\ref{tab:training_strategy_ablation}, the multi-stage strategy consistently outperforms the mixed-data alternative across benchmarks.
\begin{wraptable}{r}{0.55\textwidth}
\vspace{-10pt}
\centering
\caption{\textbf{Ablation on training strategy.} 
Multi-stage training outperforms one-stage mixed training.}
\vspace{-5pt}
\label{tab:training_strategy_ablation}
\renewcommand{\arraystretch}{1.0}
\setlength{\tabcolsep}{4pt}
\begin{tabular}{lccc}
\toprule
\textbf{Strategy} & \textbf{MuirBench} & \textbf{BLINK} & \textbf{Mantis} \\
\midrule
LLaVA-OV & 42.5 & 51.1 & 60.4 \\
One-Stage & 46.2 & 55.1 & 65.4 \\
\rowcolor{cyan!10}
Multi-Stage & \textbf{48.5} & \textbf{55.9} & \textbf{69.6} \\
\bottomrule
\end{tabular}
\vspace{-10pt}
\end{wraptable}
We attribute this improvement to the sequential learning structure. In the first stage, the model learns to capture global context and image-level coherence through structured, sequence-aligned supervision. Once this foundation is established, the second stage refines the model’s ability to attend to local, fine-grained visual cues via region-specific preference optimization. In contrast, the one-stage strategy may dilute the model’s focus by simultaneously exposing it to competing global and local objectives, making optimization less efficient.
These results suggest that decoupling context modeling and fine-grained grounding into separate stages can better guide the model toward hierarchical visual reasoning.

\section{More Observations}
\begin{figure*}[htbp]
  \centering
    \includegraphics[width=1\textwidth]{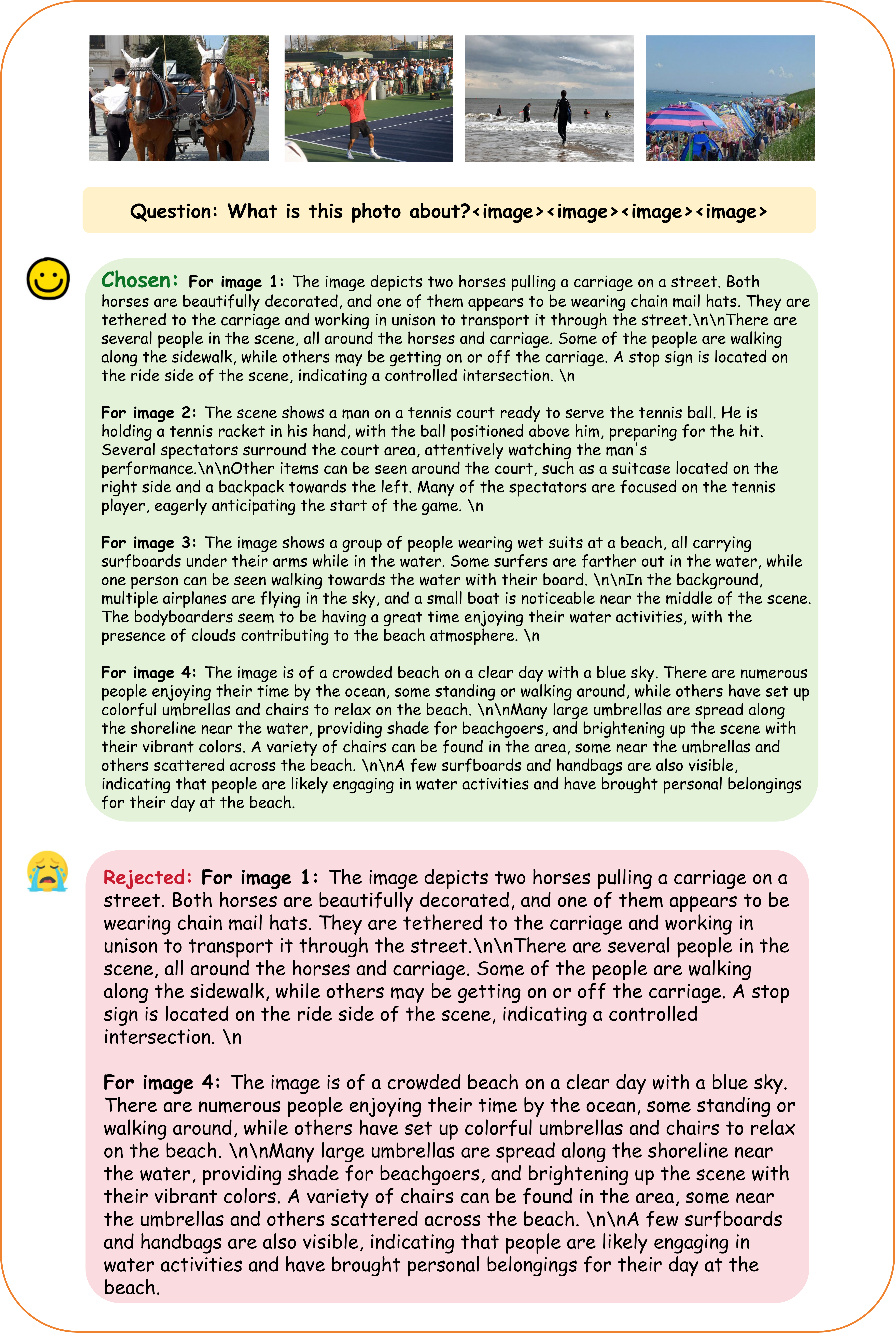}
\caption{Example of Context-Level preference pair with truncation perturbation~(Context Omission).}
  \label{fig33}
\end{figure*}

\begin{figure*}[htbp]
  \centering
    \includegraphics[width=1\textwidth]{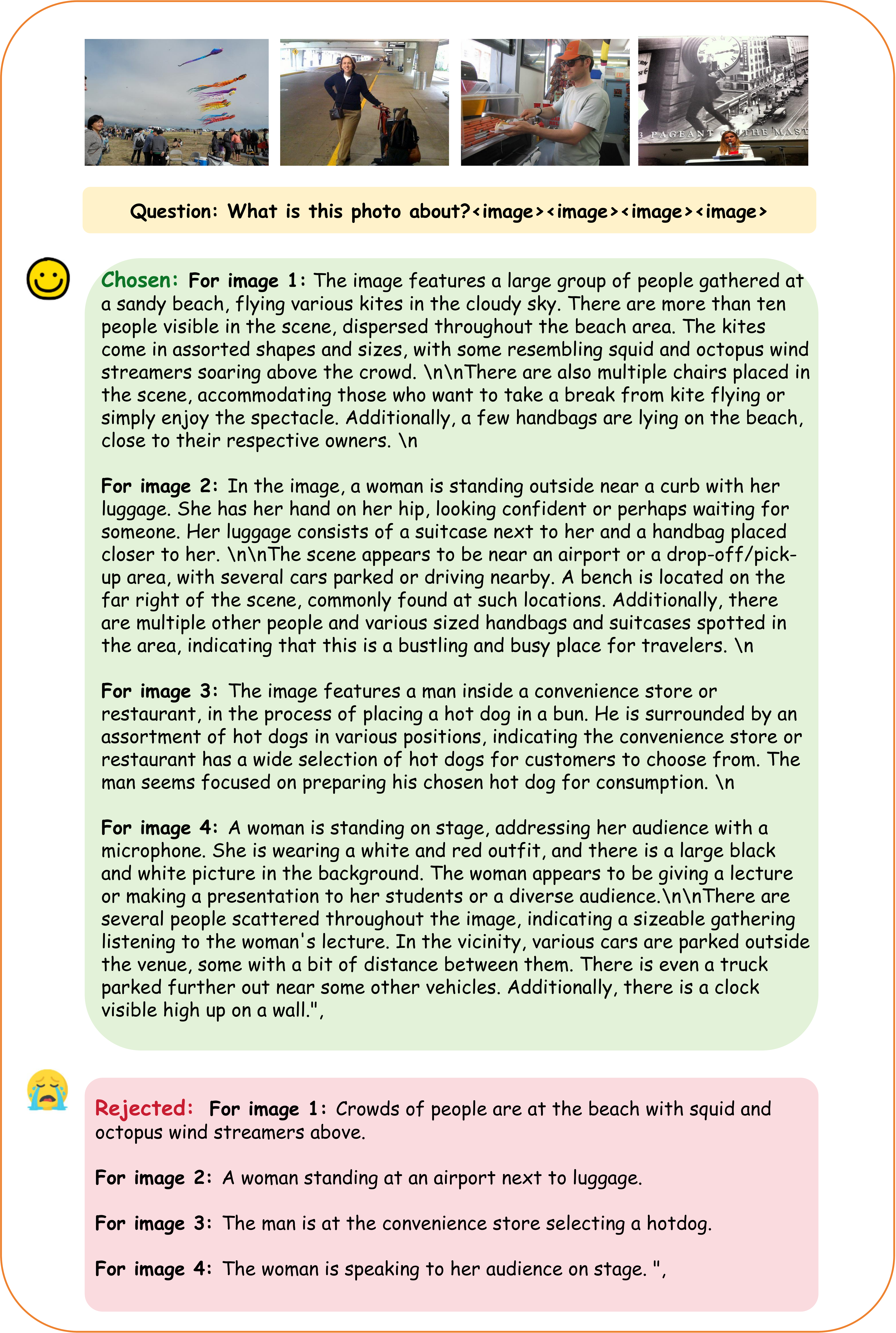}
\caption{Example of Context-Level preference pair with caption shortening perturbation~(Context Omission).}
  \label{fig55}
\end{figure*}

\begin{figure*}[htbp]
  \centering
    \includegraphics[width=1\textwidth]{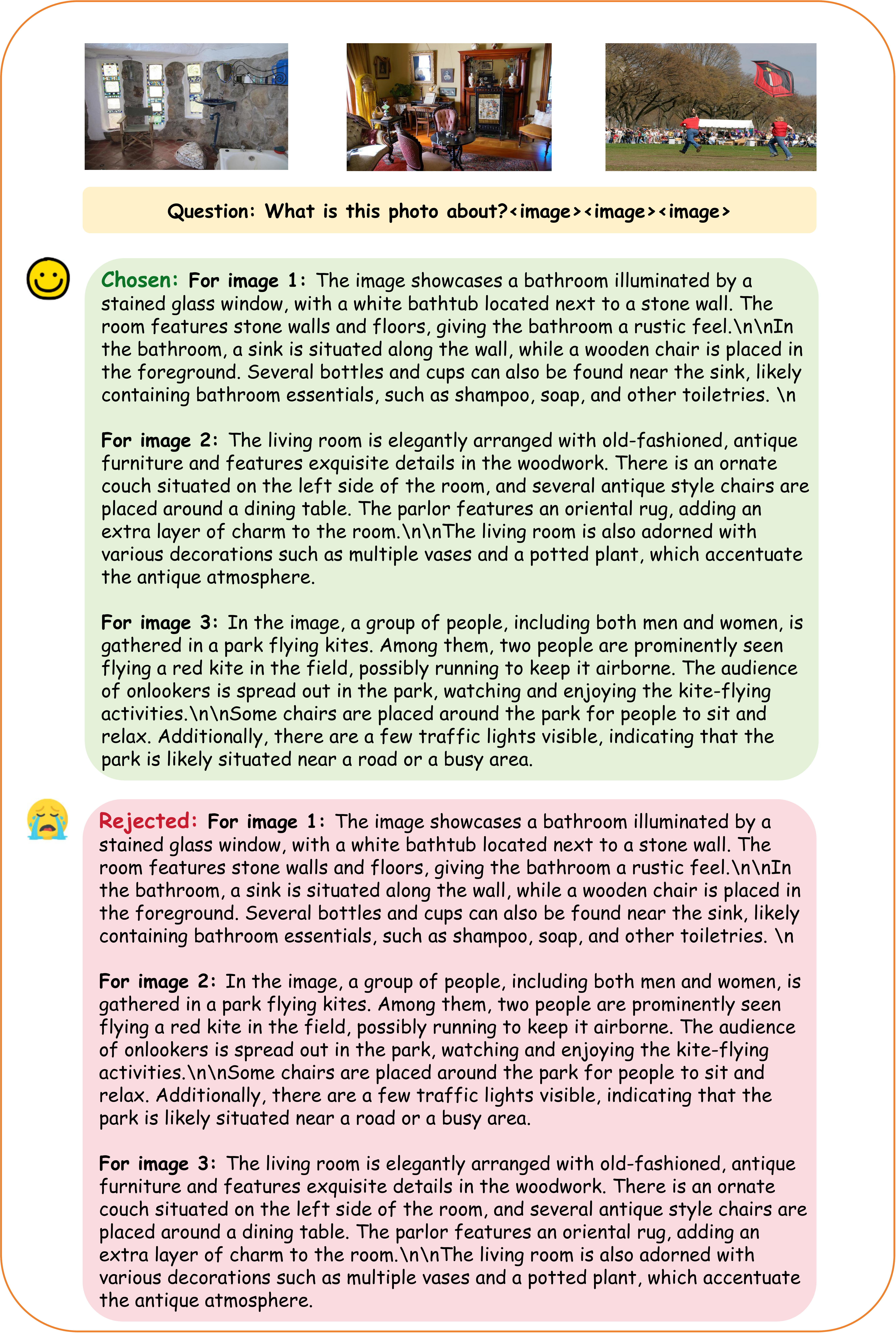}
\caption{Example of Context-Level preference pair with swapping perturbation (Context Conflation).}
  \label{fig44}
\end{figure*}

As illustrated in Fig.~\ref{fig33},~\ref{fig44},~\ref{fig55},~\ref{fig66}, and~\ref{fig77}, we present additional qualitative examples of the constructed preference pairs used in our CcDPO training. These examples cover a range of perturbation types across both Context-Level and Needle-Level supervision, each designed to target specific failure modes in multi-image reasoning.

\noindent \textbf{Fig.~\ref{fig33}} shows a context-level \textit{complete content omission} scenario, where the rejected response omits part of the image sequence. This simulates a context omission error, encouraging the model to favor complete and coherent descriptions across multiple images.

\noindent \textbf{Fig.~\ref{fig55}} demonstrates a context-level \textit{partial content omission} scenario, where the rejected response uses overly brief or generic captions. Such examples train the model to prefer rich, fine-grained visual descriptions that capture meaningful details.

\noindent \textbf{Fig.~\ref{fig44}} presents a case of \textit{caption swapping}, where descriptions are mismatched with image indices. This context conflation perturbation helps the model learn to correctly associate content with its corresponding image.

\begin{figure*}[htbp]
  \centering
    \includegraphics[width=1\textwidth]{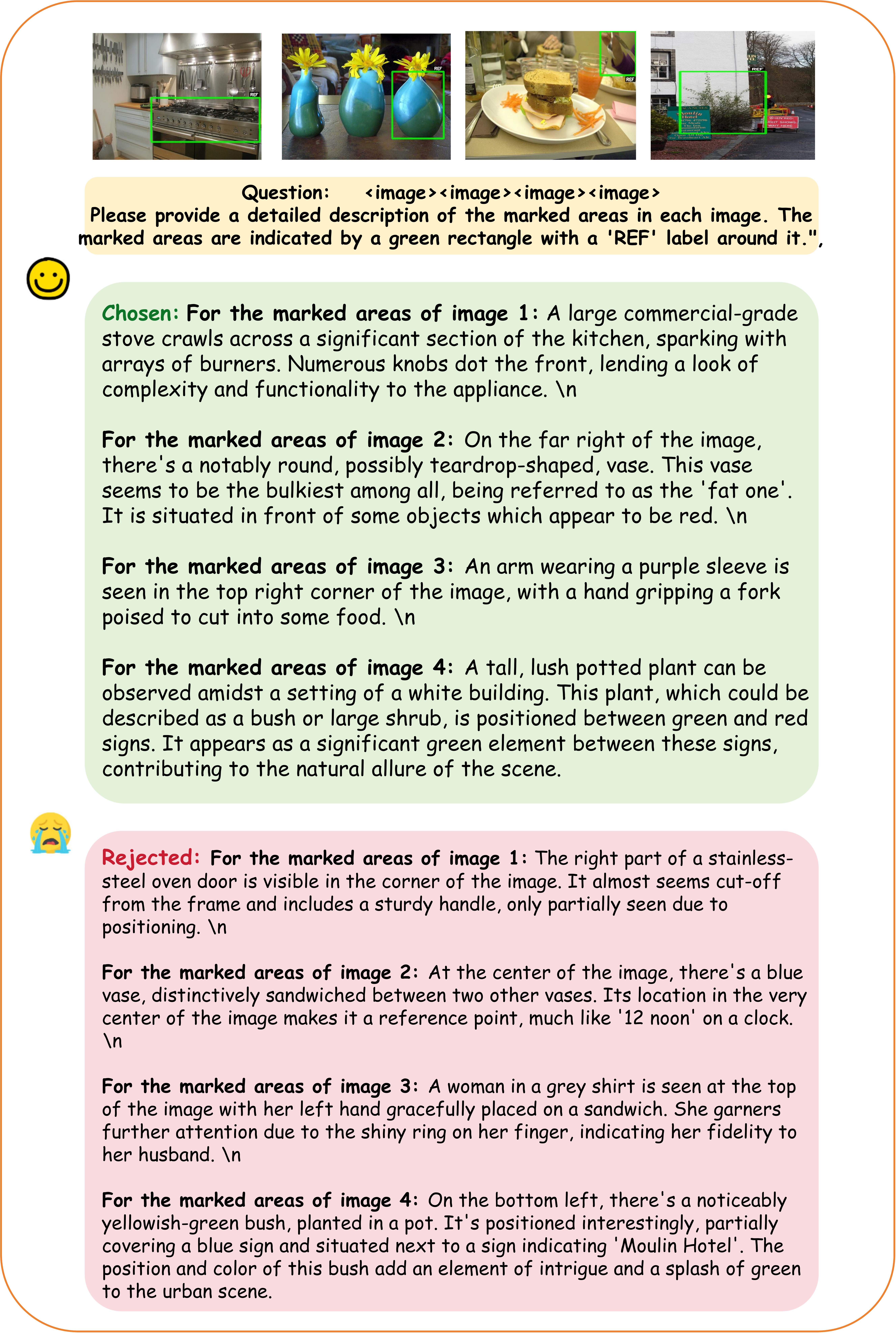}
\caption{Example of Needle-Level preference pair with bbox region mismatches perturbation (Detail
Misinterpret).}
  \label{fig88}
\end{figure*}

\begin{figure*}[htbp]
  \centering
    \includegraphics[width=1\textwidth]{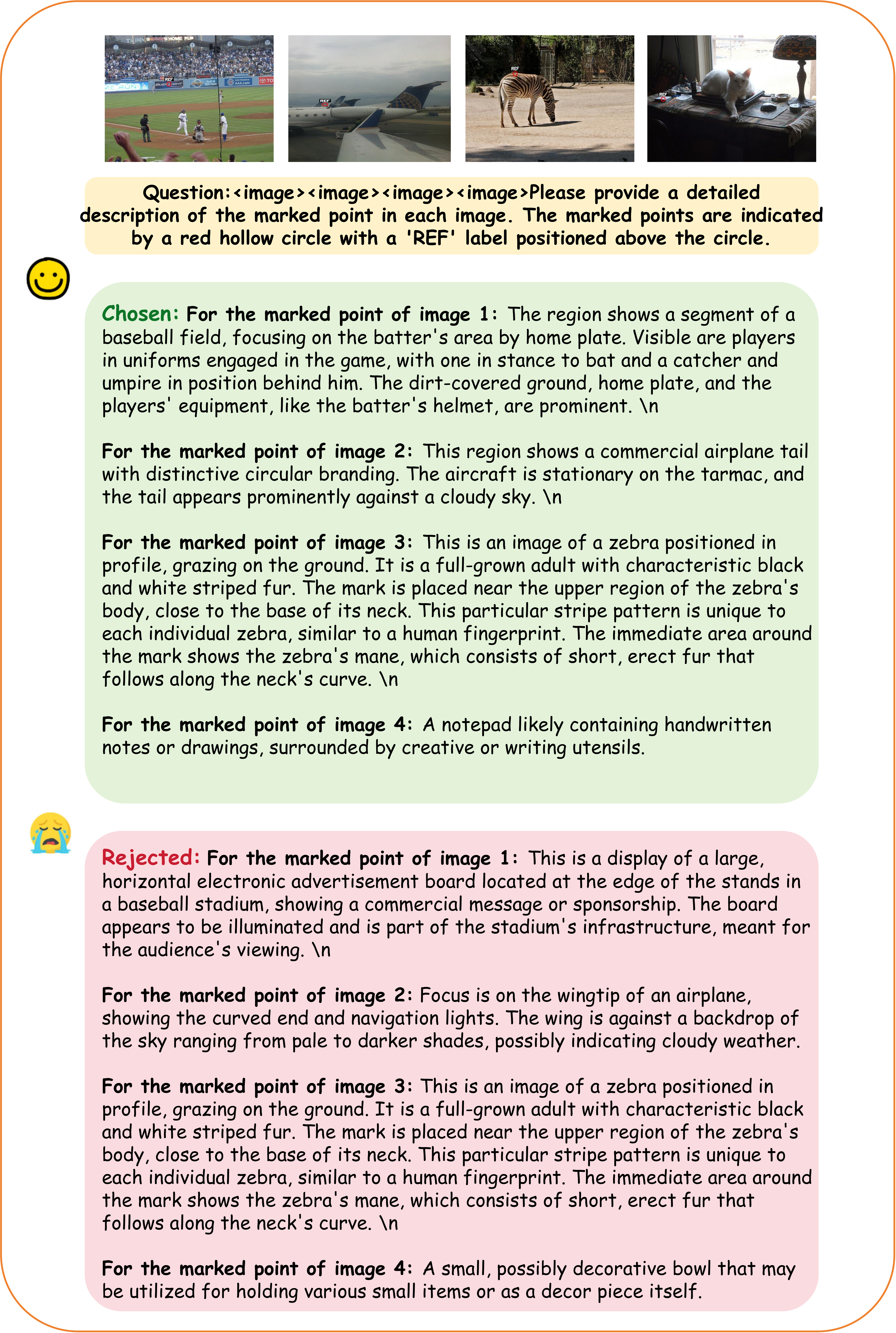}
\caption{Example of Needle-Level preference pair with point region mismatches perturbation (Detail Misinterpret).}
  \label{fig66}
\end{figure*}

\begin{figure*}[htbp]
  \centering
    \includegraphics[width=1\textwidth]{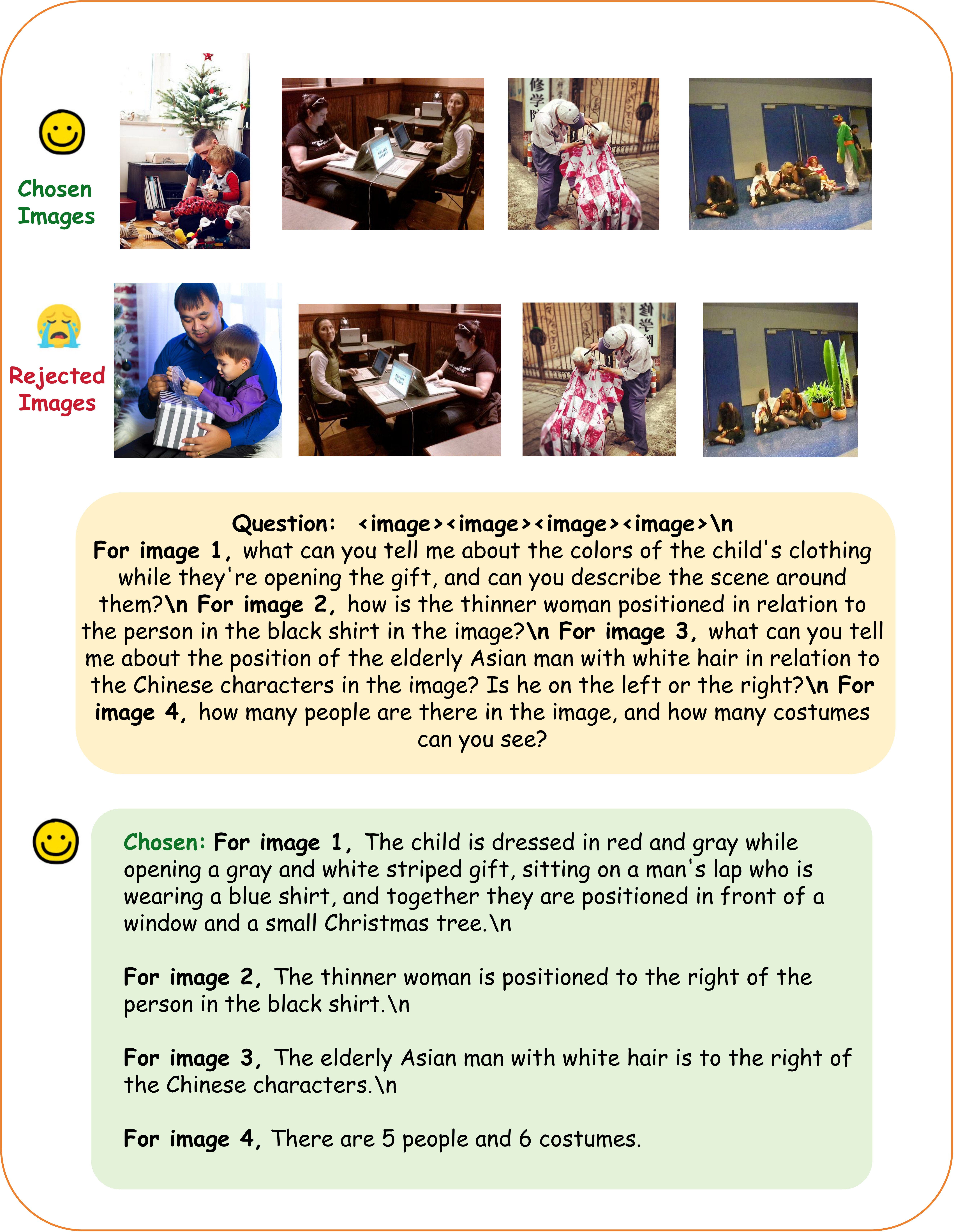}
\caption{Example of Needle-Level preference pair with image contrastive perturbation (Detail
Misinterpret).}
  \label{fig77}
\end{figure*}

\noindent \textbf{Fig.~\ref{fig88} and Fig.~\ref{fig66}} move to the Needle-Level and show a case where region-specific (bbox or point) captions are mismatched. This encourages the model to distinguish between accurate and inaccurate local grounding within a given image.

\noindent \textbf{Fig.~\ref{fig77}} illustrates the use of contrastive image inputs for the same caption, where one image visually aligns with the text while the other introduces a subtle contradiction. This visual preference pair sharpens the model’s ability to resolve ambiguous references through visual evidence alone.

Together, these examples highlight how different types of preference signals are systematically constructed to improve the model’s ability to reason over sequences, resolve visual references, and discriminate between subtle semantic mismatches. By explicitly encoding both global and local reasoning objectives into the supervision data, CcDPO promotes more robust and interpretable multi-image alignment.

\end{document}